%% file: main.tex
\documentclass[letterpaper]{article} % DO NOT CHANGE THIS
\usepackage{aaai24}  % DO NOT CHANGE THIS
\usepackage{times}  % DO NOT CHANGE THIS
\usepackage{helvet}  % DO NOT CHANGE THIS
\usepackage{courier}  % DO NOT CHANGE THIS
\usepackage[hyphens]{url}  % DO NOT CHANGE THIS
\usepackage{graphicx} % DO NOT CHANGE THIS
\usepackage{natbib}  % DO NOT CHANGE THIS AND DO NOT ADD ANY OPTIONS TO IT
\usepackage{caption} % DO NOT CHANGE THIS AND DO NOT ADD ANY OPTIONS TO IT
\usepackage{algorithm}
\usepackage{algorithmic}

\usepackage{newfloat}
\usepackage{listings}
\DeclareCaptionStyle{ruled}{labelfont=normalfont,labelsep=colon,strut=off} % DO NOT CHANGE THIS
\floatstyle{ruled}
\newfloat{listing}{tb}{lst}{}
\floatname{listing}{Listing}
\title{Recall-Oriented Continual Learning with Generative Adversarial Meta-Model}
\author{
    Haneol Kang,
    Dong-Wan Choi\thanks{Corresponding Author.}
}
\affiliations{
    Department of Computer Science and Engineering, Inha University, South Korea\\
    haneol0415@gmail.com, dchoi@inha.ac.kr
}

\usepackage{subfigure}
\usepackage{xspace}
\usepackage{amsmath}
\usepackage{amssymb}
\usepackage{multirow}
\usepackage{makecell}
\usepackage{array}
\usepackage{xcolor}  % should be deleted when submission
\newcolumntype{L}[1]{>{\raggedright\let\newline\\\arraybackslash\hspace{0pt}}m{#1}}
\newcolumntype{C}[1]{>{\centering\let\newline\\\arraybackslash\hspace{0pt}}m{#1}}
\newcolumntype{R}[1]{>{\raggedleft\let\newline\\\arraybackslash\hspace{0pt}}m{#1}}

\newcommand{\eat}[1]{}
\newcommand{\smalltitle}[1]{ \vspace{1mm}{\noindent\textbf{#1.}\hspace{1mm}}}

\newcommand*\soline[1]{%
  \vbox{%
    \hrule height 0.5pt%                  % Line above with certain width
    \kern0.25ex%                          % Distance between line and content
    \hbox{%
      \kern-0.3em%                        % Distance between content and left side of box, negative values for lines shorter than content
      \ifmmode#1\else\ensuremath{#1}\fi%  % The content, typeset in dependence of mode
      \kern-0.3em%                        % Distance between content and left side of box, negative values for lines shorter than content
    }% end of hbox
  }% end of vbox
}

\newcommand*\loline[1]{%
  \vbox{%
    \hrule height 0.5pt%                  % Line above with certain width
    \kern0.25ex%                          % Distance between line and content
    \hbox{%
      \kern-0.2em%                        % Distance between content and left side of box, negative values for lines shorter than content
      \ifmmode#1\else\ensuremath{#1}\fi%  % The content, typeset in dependence of mode
      \kern-0.2em%                        % Distance between content and left side of box, negative values for lines shorter than content
    }% end of hbox
  }% end of vbox
}

\newcommand{\alglinelabel}{%
  \addtocounter{ALC@line}{-1}% Reduce line counter by 1
  \refstepcounter{ALC@line}% Increment line counter with reference capability
  \label% Regular \label
}

\def\L{\mathcal{L}}
\def\S{\mathcal{S}}
\def\V{\mathcal{V}}

\def\D{\mathcal{D}}
\def\B{\mathcal{B}}
\def\G{\mathcal{G}}
\def\R{\mathcal{R}}
\DeclareMathOperator*{\argmin}{arg\,min}

\def\ourmodel{GAMM\xspace}

\begin{document}

\maketitle

\begin{abstract}
The \textit{stability-plasticity dilemma} is a major challenge in continual learning, as it involves balancing the conflicting objectives of maintaining performance on previous tasks while learning new tasks. In this paper, we propose the \textit{recall-oriented continual learning framework} to address this challenge. Inspired by the human brain's ability to separate the mechanisms responsible for stability and plasticity, our framework consists of a two-level architecture where an inference network effectively acquires new knowledge and a generative network recalls past knowledge when necessary. In particular, to maximize the stability of past knowledge, we investigate the complexity of knowledge depending on different representations, and thereby introducing \textit{generative adversarial meta-model} (GAMM) that incrementally learns task-specific parameters instead of input data samples of the task. Through our experiments, we show that our framework not only effectively learns new knowledge without any disruption but also achieves high stability of previous knowledge in both task-aware and task-agnostic learning scenarios. Our code is available at: \url{https://github.com/bigdata-inha/recall-oriented-cl-framework}.
\end{abstract}

%--------------------------------------------------------------------------------------

\section{Introduction} \label{sec:intro}
\begin{figure}[t]
  \centering 
   \includegraphics[width=0.8\columnwidth]{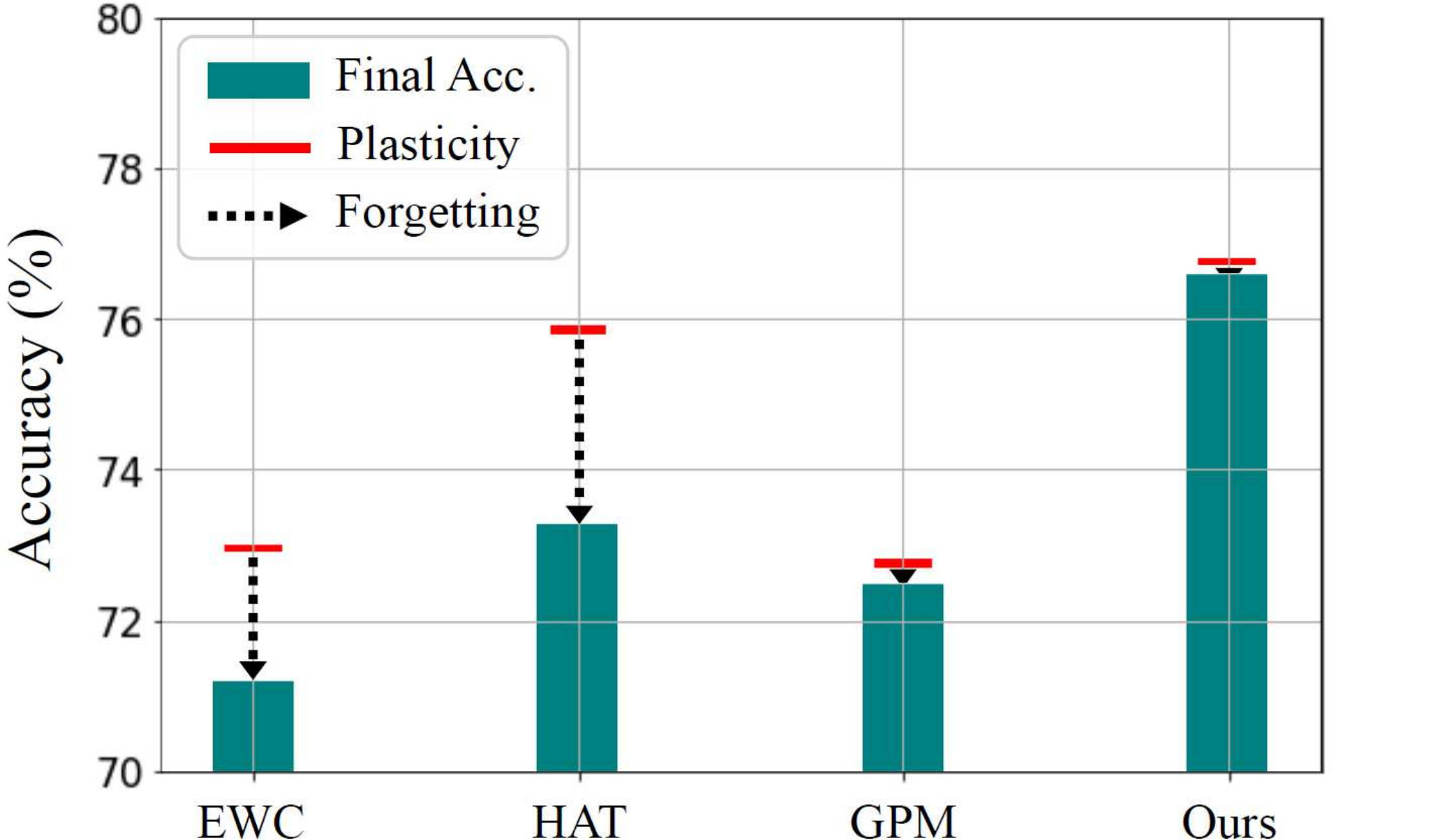}
  \caption{Trade-off between stability and plasticity in various CL baseline approaches on Split CIFAR-100.}
  \label{fig:intro_fig:a}
\end{figure}

The ability to continuously acquire new knowledge without forgetting previously learned information is a major challenge in continual learning (CL) over deep neural networks. This challenge, often referred to as the \textit{stability-plasticity dilemma} \cite{ParisiKPKW19}, requires balancing two opposing objectives: preserving past knowledge (stability) while adapting to new tasks (plasticity). On the one hand, if we focus on stability by preventing the model's state from undergoing significant changes, the model's plasticity to acquire new information is impaired. On the other hand, making the model highly adaptable to incoming tasks leads to stability degradation on previous tasks, known as \textit{catastrophic forgetting} \cite{mccloskey1989catastrophic}. Despite considerable progress of recent studies in CL, the trade-off between learning new tasks and maintaining old knowledge remains somewhat inevitable. As shown in Figure~\ref{fig:intro_fig:a}, methods with high plasticity experience significant forgetting, while those with less forgetting lack the ability to adapt to new tasks.

In contrast, the human brain gracefully resolves the dilemma by separating yet complementing the parts responsible for stability and plasticity. More specifically, working memory (\textit{a.k.a.} short-term memory) focuses on processing new information, whereas long-term memory is responsible for retaining and consolidating the information that is not currently used but may be needed in the future \cite{mcclelland1995there,o2014complementary}. More interestingly, when we recall past knowledge, it is known that our brain creates something new in a generative manner from long-term memory, rather than simply retrieving something that explicitly exists in the brain \cite{schacter2007cognitive}.

Existing brain-inspired CL methods also consider a certain level of separation of new knowledge from previous knowledge, but mostly focus on the \textit{rehearsal} process of human memory. This has led to the development of techniques like \textit{memory replay} \cite{lopez2017gradient,rebuffi2017icarl} and \textit{generative replay} \cite{KemkerK18,ShinLKK17}. Memory replay stores representative old samples in an extra buffer, while generative replay trains a generative model that can generate pseudo-samples of previous tasks. Then, in both techniques, previous samples or pseudo-samples are jointly trained with new data in a single neural network. However, due to limited memory and the complexity of underlying data distribution, neither approach can fully capture the entire knowledge of previous tasks, which results in less satisfactory performance.

\begin{figure}[t]
  \centering 
  \includegraphics[width=0.88\columnwidth]{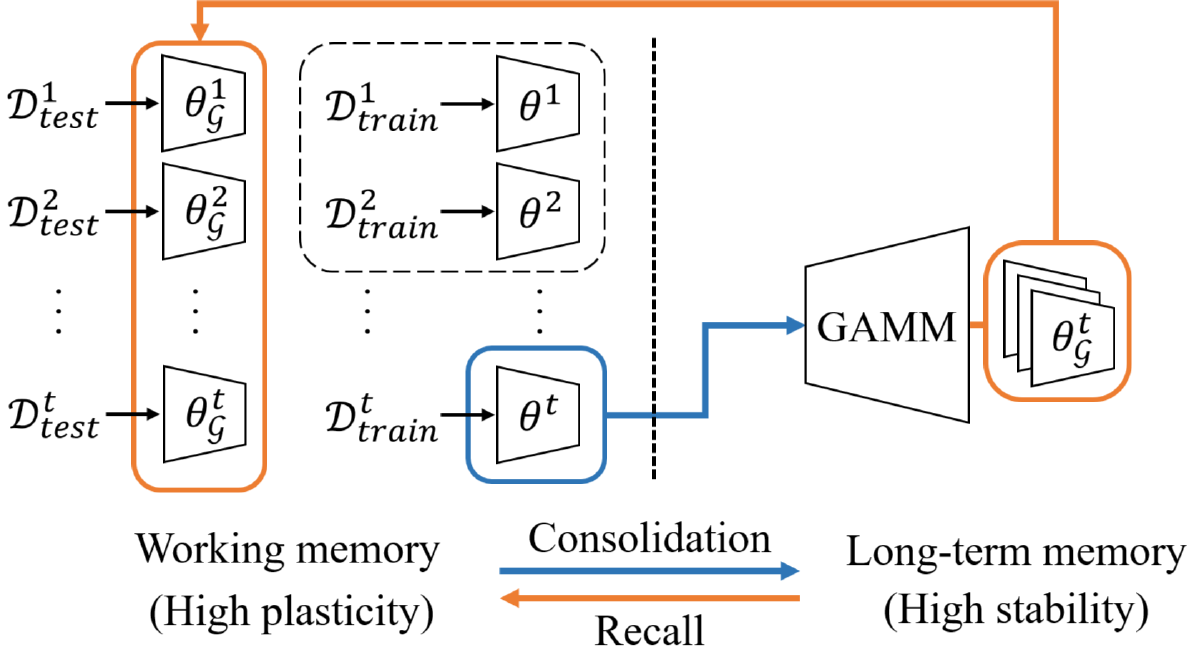}
  \caption{Separation of working memory and long-term memory in our framework.}
  \label{fig:intro_fig:b}
\end{figure}

In this paper, we propose the \textit{recall-oriented CL framework}, where we divide learning process into two different networks corresponding to the human brain's working memory and long-term memory, as illustrated in Figure~\ref{fig:intro_fig:b}. Firstly, a typical neural network takes the role of working memory, and freely learns new knowledge without any disruption to achieve pure plasticity. Secondly, inspired by the generative way of the human's recall process \cite{schacter2007cognitive}, we introduce \textit{generative adversarial meta-model} (\ourmodel) to be our counterpart of long-term memory. \ourmodel is a generative model that trains over model parameters rather than data samples, which thus can directly \textit{recall} a previous task-specific model at inference time. The \ourmodel's approach stems from our analysis on the complexity of knowledge based on different representations, assessed through two metrics: separability and volume. Our analysis reveals that the complexity of raw data is significantly higher than that of the corresponding learned parameters, implying that a trained model itself is a better form of knowledge to be accumulated in long-term memory than its training data distribution.

The major challenge in our framework is how to train \ourmodel in an efficient and scalable manner. One issue is the need of numerous versions of learned parameters for each task, as a typical generative model takes many samples to be trained. A possible approach can be training the task-specific model multiple times with different initialized parameters or different subsets of training data \cite{JosephB20}. However, this approach of redundant training is not only inefficient \cite{RatzlaffL19}, but also less effective in acquiring the knowledge of new tasks, often requiring fine-tuning at inference time \cite{JosephB20}. Our strategy is to use \textit{Bayesian neural network} (BNN) \cite{blundell2015weight, MaddoxIGVW19} as a task-specific model without redundant training, and \ourmodel takes a sufficient number of different models for the same task from the task-specific BNN. In terms of scalability, it is also important for \ourmodel not to use too large memory space to accumulate the learned knowledge. To this end, we observe that each individual task may not require a large capacity for plasticity, and thus we keep each task-specific model as lightweight as possible. This allows \ourmodel to incorporate many task-specific parameters within its limited capacity.

In our experiments, we show that our recall-oriented CL framework nicely addresses the stability-plasticity dilemma with a small amount of memory. Compared to existing replay-based methods, our framework achieves the best performance while mostly consuming less memory space in both task-aware and task-agnostic CL scenarios. Also, our two-level architecture is observed to be highly effective at plasticity, and therefore outperforms the existing CL baseline approaches.

%--------------------------------------------------------------------------------------

\section{Related Works} \label{sec:related}
\smalltitle{Continual Learning for Less Forgetting}
Existing methods of continual learning in neural networks can be grouped into three main categories: \textit{regularization-based}, \textit{architecture-based}, and \textit{replay-based methods}. Regularization-based methods \cite{AhnCLM19,EbrahimiEDR20,kirkpatrick2017overcoming,ZenkePG17} more focus on stability by penalizing updates to important parameters, resulting in lower plasticity. On the other hand, architecture-based methods aim to adaptively utilize different parts of the network for each task by masking and attentions \cite{mallya2018packnet,SerraSMK18}, or by dynamically expanding the network for incoming tasks \cite{ostapenko2019learning,YoonYLH18}. Replay-based methods attempt to mimic the human rehearsal mechanism by storing previous knowledge, typically in the form of exemplars \cite{lopez2017gradient,rebuffi2017icarl,SahaG021,BuzzegaBPAC20,LyuW0YH021} or as a generative model \cite{KemkerK18,ShinLKK17,van2020brain}. Both suffer from the stability-plasticity dilemma due to limited memory and complex data distributions.

\smalltitle{Parameter Generation}
A \textit{parameter-generation} model (\textit{a.k.a.} \textit{meta-model}) is a type of model that is trained to generate parameters, rather than data samples. One popular framework is \textit{hypernetwork} \cite{HaDL17}, in which a target network is created by a deterministic hypernetwork and is not independently trained. Hypernetworks and their variants \cite{krueger2017bayesian, RatzlaffL19} have been applied to various tasks including few-shot learning \cite{Zhmoginov0V22}, uncertainty estimation \cite{RatzlaffL19}, and continual learning \cite{OswaldHSG20}. Our parameter-generation scheme is different from hypernetworks in that we train a \textit{generative adversarial network} (GAN) \cite{NIPS2014_5ca3e9b1} over multiple task-specific models, each separately trained. BNNs \cite{blundell2015weight, MaddoxIGVW19} can also be seen as parameter-generation models in that they treat network parameters as random variables and learn their distribution. Our framework trains \ourmodel by taking multiple parameters sampled from a single task-specific BNN. Although APD \cite{WangVLGGZ18} also suggests training a GAN based on BNNs, they focus on how to replace MCMC-based BNNs requiring excessive training with a corresponding GAN, not considering any scenario where multiple tasks must be sequentially learned within a single GAN.

\smalltitle{Continual Learning with Parameter Generation}
A few studies have investigated continual learning using parameter-generation models. Most existing methods rely on hypernetworks, where knowledge for each new task is not separately learned but directly consolidated into the meta-model \cite{HenningCDOTEKGS21,OswaldHSG20}. This one-phase training scheme seems excellent at retaining stability on previous tasks, achieving almost zero-forgetting, but it sacrifices plasticity as demonstrated by our experimental results. In addition, hypernetwork-based methods exhibit limited performance particularly when task identities are not given, likely due to the failure to adequately separate task-specific knowledge in the hypernetwork. Without using hypernetworks, our framework is similar to MERLIN \cite{JosephB20} in that both employ non-deterministic meta-models, a \textit{variational autoencoder} (VAE) \cite{kingma2013auto} for MERLIN and a GAN for ours. However, MERLIN is designed for online learning scenarios and thus its meta-model is trained to generate rather small-sized networks (\textit{e.g.,} an order of magnitude smaller than those of both hypernetworks and ours). This consequently requires fine-tuning at inference time with an extra memory buffer, as their goal is to generate well-initialized models. In contrast, our framework is capable of generating high-end models that are ready to use without fine-tuning.

%--------------------------------------------------------------------------------------

\section{Analysis on Representation Complexity} \label{sec:analysis}
In a two-level CL approach, such as ours, where old and new knowledge are somehow separately maintained, a crucial question is what would be the best representation to preserve the learned knowledge. For example, in replay-based approaches, the previous knowledge is not only embedded in the inference model being trained but also separately stored in a buffer of exemplars or in a generative model, both of which however suffer from the stability-plasticity dilemma as mentioned earlier. To examine the above question, this section investigates how to quantify the complexity of knowledge depending on different representations, and introduces two metrics applicable to heterogeneous representations in different structures and dimensions. Finally, we specifically analyze and compare the complexity of raw data samples, feature vectors, and learned parameters.

\subsection{Representation Complexity Metrics}

In order to quantify the complexity of representations for a given training dataset, we take into account the following two perspectives, namely \textit{volume} ($\mathcal{V}$) and \textit{separability} ($\mathcal{S}$). 

\smalltitle{Volume} The volume is to measure the diversity of knowledge representations. Thus, a larger volume indicates the need for a larger capacity to learn diverse cases. To this end, we suggest measuring how widely representation points are distributed within each local group (\textit{e.g.,} class), not across the entire representation space. This is based on the intuition that points belonging to different groups are easier to discriminate and thus simpler to learn when they are farther apart, which however undesirably lead to a wider distribution overall. Thus, in terms of the difficulty of learning, diversity needs to be measured locally, rather than globally across the entire space. In addition, the volume should not be dependent on dimensionality, and therefore dimensionality reduction is essential to compare representations with different dimensions. Based on these criteria, we define the volume ($\mathcal{V}$) as follows:
\begin{equation}
    \mathcal{V} = \sum_{c=1}^C \det(\Sigma_c)^{1/2} = \sum_{c=1}^C \prod_{i=1}^d {\lambda_i^c}^{1/2} \;,
\end{equation}
where $C$ is the number of local groups (e.g., number of classes), $\Sigma_c$ is the covariance matrix for group $c \in [1, C]$, and $\lambda_i^c$ is its $i$-th eigenvalue. Thus, $\mathcal{V}$ is proportional to the square root of the determinant of its covariance matrix, which is the product of eigenvalues $\lambda_i^c$. As $d$ is equally applied to representations with different dimensions, the volume is independent of dimensionality. It is also noteworthy that the more the local groups there are in the representation space, the larger the corresponding volume is, and thereby capturing the global diversity as well as the local diversity.

\smalltitle{Separability} The separability refers to how well representations are separated across different groups (\textit{e.g.,} classes) while being clustered within each group. Intuitively, well-separated representations lead to low entropy, making them easier to learn. Similar to Fisher's discriminant ratio \cite{fisher1936use} and Davies-Bouldin Index \cite{DaviesB79}, we define the separability ($\mathcal{S}$) as the ratio of \textit{between-group} variance to total variance as follows:
\begin{equation}
    \mathcal{S} = \frac{\text{Var}[\mathbb{E}(x|c)]}{\text{Var}(x)} = \frac{\text{Var}[\mathbb{E}(x|c)]}{\text{Var}[\mathbb{E}(x|c)] + \mathbb{E}[\text{Var}(x|c)]}\;,
\end{equation}
where the total variance $\text{Var}(x)$ is decomposed into \textit{between-group} variance $\text{Var}[\mathbb{E}(x|c)]$ and \textit{within-group} variance $\mathbb{E}[\text{Var}(x|c)]$. Thus, $\mathcal{S}$ is minimized (\textit{i.e.,} $0$) when all the groups are not separated at all to the point that their means are exactly the same, making $\text{Var}[\mathbb{E}(x|c)] = 0$, whereas $\mathcal{S}$ is maximized (\textit{i.e.,} $1$) when each group has the maximum cohesion with $\mathbb{E}[\text{Var}(x|c)] = 0$.

\subsection{Analytic Results}

\smalltitle{Setting}
We evaluate the complexity of three types of representations: input-level, feature-level, and parameter-level representations, using the MNIST \cite{LeCunBBH98}, SVHN \cite{netzer2011reading}, and CIFAR-10 \cite{krizhevsky2009learning} datasets, equally consisting of 10 classes. Input-level representations are training images themselves, while feature-level representations are their feature vectors from a model trained on the images. Finally, for the parameter-level representation, we train a Bayesian neural network (BNN) and take model samples from the BNN as many as the number of input images. Each sampled model is then split into the same number of chunks as the number of classes (\textit{i.e.,} 10 for all three datasets), and these chunks and classes are regarded as local groups when computing the volume and separability. ResNet-32 \cite{HeZRS16} is commonly used for the model architecture throughout the analytic process.

\begin{figure}[t!]
\centering
\includegraphics[width=0.9\columnwidth]{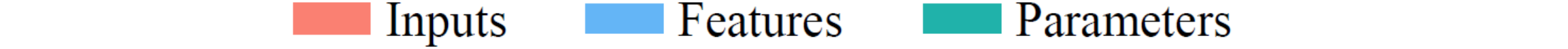}
\subfigure[\label{fig:mnist_sv}MNIST]{\hspace{-1mm}\includegraphics[width=0.33\columnwidth]{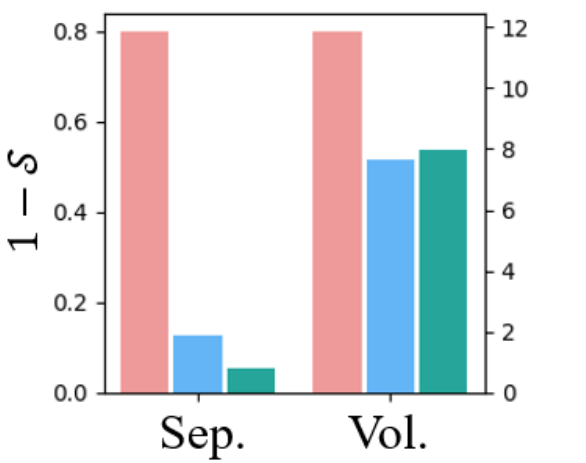}}
\subfigure[\label{fig:svhn_sv}SVHN]{\hspace{-1mm}\includegraphics[width=0.33\columnwidth]{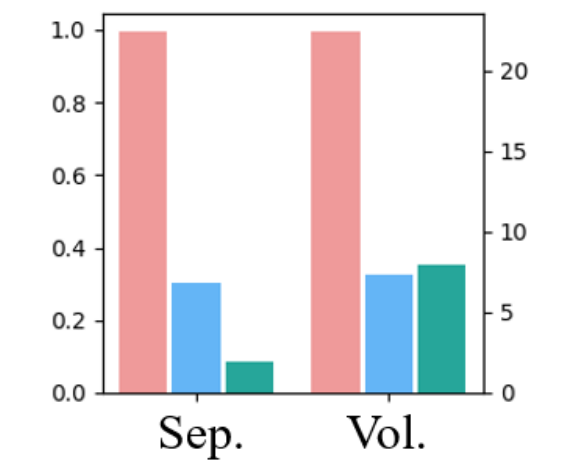}}
\subfigure[\label{fig:cifar_sv}CIFAR-10]{\hspace{-1mm}\includegraphics[width=0.33\columnwidth]{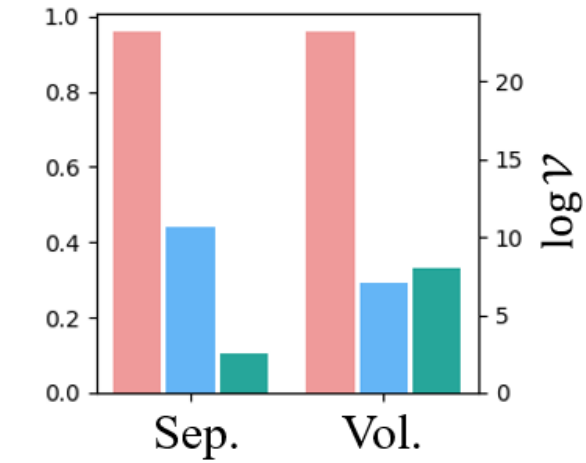}}
\caption{Separability and volume of input images, feature vectors, and parameter chunks.}\label{fig:sep_vol}
\end{figure}

\smalltitle{Results}
Figure \ref{fig:sep_vol} presents the separability and volume of three types of representations using different datasets, where the left y-axis represents values of $1-\S$, while the right y-axis depicts the values of $\V$ on a logarithmic scale. Thus, smaller bar heights indicate lower complexity in terms of separability and volume. As expected to some extent, input-level representations exhibit bars much longer than those of features and parameters. This implies that the representation of raw data is too complex either to be adequately covered by a few exemplars or a generative model. As a result, both memory replay and generative replay encounter challenges of preserving knowledge when dealing with complex datasets. While the feature-level representation demonstrates significantly reduced complexity compared to input images, it still proves to be more complex in terms of separability than parameter chunks. This observation is noteworthy since the dimensionality of parameter chunks is typically much higher than that of feature vectors. 

In summary, despite their high dimensionality, learned parameters stand out as the most compact representation capable of efficiently capturing the underlying data distribution. Based on this analysis, our framework suggests the generative replay of learned parameters instead of using data samples or feature vectors.

%--------------------------------------------------------------------------------------

\section{Methodology} \label{sec:method}
\begin{figure*}[t]
  \centering 
    \includegraphics[width=0.8\textwidth]{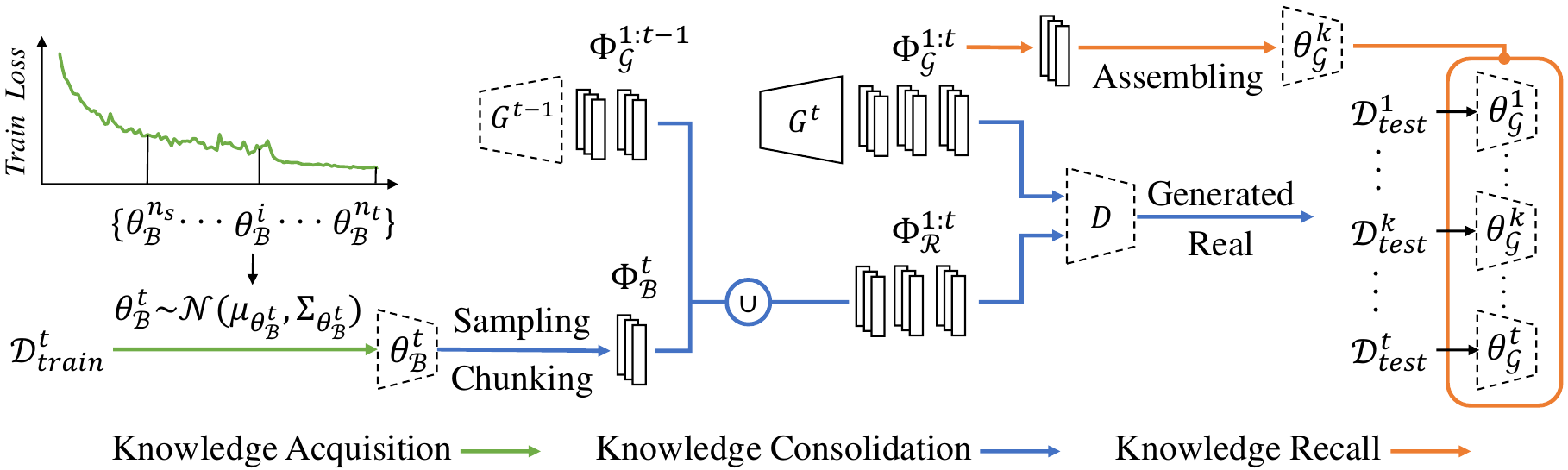}
  \caption{Illustration of three major steps in our recall-oriented continual leaning framework.}
  \label{fig:main_illust}
\end{figure*}
In this section, we first formulate the problem of continual learning (CL), and then present our proposed recall-oriented continual learning framework.

\smalltitle{Problem Statement} In this paper, we follow the standard task-incremental learning (TIL) scenario, where a sequence of $T$ tasks needs to be incrementally learned and each task $t$ for $t \in [1, T]$ is associated with its training data $\D^{\,t}_{train}$ and test data $\D^{\,t}_{test}$. The goal of TIL is to incrementally train a model that can make precise prediction for any task previously learned. In a typical TIL setting, we consider a task-aware scenario, where the task identity is explicitly given to the model at inference time as well as training time. Although our framework basically aims to deal with this task-aware version of TIL, it is also experimentally observed to be effective in a task-agnostic scenario without task-identifiers at inference time.

\smalltitle{Proposed Framework}
In order to achieve both high plasticity (ability to learn new knowledge) and high stability (ability to retain previous knowledge), we propose the recall-oriented CL framework that consists of three major steps, namely \textit{knowledge acquisition}, \textit{knowledge consolidation}, and \textit{knowledge recall}. The overall process of the framework is illustrated in Figure \ref{fig:main_illust}. In the knowledge acquisition step, knowledge of incoming task is acquired in the form of model parameters to be consolidated into the meta-model (\textit{i.e.,} \ourmodel). More specifically, we train a lightweight BNN, which corresponds to working memory, without any attempts to preserve the previous knowledge, which thus secures pure plasticity. The knowledge of working memory is then merged into \ourmodel that corresponds to long-term memory in the knowledge consolidation step. Motivated by our analysis in the previous section, \ourmodel is trained to generate learned parameters for the current task using the trained BNN, while trying not to forget model parameters for the past tasks through generative replay. Finally, at inference time, \ourmodel \emph{recalls} a \emph{synthetic} model for each task, which is expected to achieve the performance comparable to that of the original task-specific model.

\subsection{Knowledge Acquisition}\label{subsec:const}
The knowledge acquisition step aims to learn new knowledge by training a model specialized to each new task, and this task-specific knowledge is later consolidated into our long-term memory, \ourmodel. In a few approaches with hypernetworks, the process of knowledge consolidation is simultaneously performed during learning each new task, with simple regularization on meta-model \cite{HenningCDOTEKGS21,OswaldHSG20}. Although these approaches have shown effectiveness at knowledge preservation in the meta-model, learning with regularization may lead to parameters less optimized for the current task, hindering plasticity. In order to ensure high plasticity, we propose a two-level approach of separating the process of learning new task from training on \ourmodel.

This separated process of knowledge acquisition implies that we cannot feed data samples directly to \ourmodel unlike hypernetworks being trained using data samples. Instead, we need multiple different models themselves trained for the same task, for \ourmodel to use them as its training data. A straightforward approach to this end is training a task-specific model multiple times with differently initialized parameters \cite{Lakshminarayanan17} or with different subsets of data samples \cite{JosephB20}, and collecting all those resulting models. However, it is prohibitively inefficient to train as many different models as necessary for \ourmodel to effectively capture the underlying parameter distribution.

\smalltitle{Task-Specific BNN}
In order to address this issue, we propose using \textit{Bayesian neural network} (BNN) as a task-specific model. Unlike a typical neural network that is trained to determine a single value for each weight, a BNN approximates the posterior distribution of weights $p(\theta|\mathcal{D})$, which allows us to sample each weight from the BNN \cite{blundell2015weight}. Therefore, it suffices to train a single task-specific BNN $\theta_{\B}^{\,t} \sim p(\theta|\mathcal{D}^{t})$, instead of training multiple models for each $t$-th task. The training of $\theta_{\B}^{\,t}$ relies on how well we represent the posterior distribution of weights. In many BNNs, the posterior distribution is approximated to be a Gaussian distribution. We also make this assumption, and hence we have: $\theta_{\B}^{\,t} \sim \mathcal{N}(\mu_{\theta_\B^t}, \Sigma_{\theta_\B^t})$. Note that this Gaussian posterior distribution on model parameters is known to be effectively captured by the training history of a single neural network \cite{IzmailovPGVW18,MaddoxIGVW19}, without the need for excessive training\footnote{The details of $\mu_{\theta_\B^t}$ and $\Sigma_{\theta_\B^t}$ are presented in the Appendix.}. Once $\theta_{\B}^{\,t}$ is well-trained, we can take multiple parameter samples from the BNN and feed them to \ourmodel, enabling \ourmodel to effectively learn the parameter distribution of the model for the task $t$.

When training a task-specific BNN, we use a lightweight model rather than one with a large number of parameters. This is based on our intuition that each individual task may not require such a high capacity to be trained effectively. Also, using a lightweight model is essential for \ourmodel to reduce its required capacity.

\subsection{Knowledge Consolidation} \label{subsec:consol}

In the knowledge consolidation step, the acquired knowledge in working memory, which is learned by a task-specific BNN, is transferred to our long-term memory, \ourmodel. During this process, it is important to avoid forgetting previously learned knowledge that has been consolidated into \ourmodel. Thus, once plasticity is achieved in knowledge acquisition, \ourmodel is responsible for accumulating task-specific knowledge with high stability. As the knowledge is acquired in the form of parameters, our goal is to incrementally train \ourmodel in a way that it can re-generate task-specific parameters for all the tasks learned so far.

We design \ourmodel to be a conditional GAN, and thus we can apply generative replay to prevent \ourmodel from forgetting the parameter distributions of the past tasks. Unlike typical generative replay aiming at input-level generation, we replay parameters for the previous tasks, and train them with parameters of the current task. As analyzed in the previous section, the parameter-level representation is much more compact than the corresponding input-level representation, thereby achieving high stability with a limited capacity.

\smalltitle{Initial Learning on \ourmodel} Let us now present in detail how to incrementally train \ourmodel by generative replay on parameter-level representations. For learning the first task with \ourmodel, we can adopt the standard way of training a GAN without replay. In typical GAN training, a generator $G$ and a discriminator $D$ are alternately trained with their conflicting objectives, namely generating samples close to the real ones and identifying those fake samples, respectively. In recent studies on GAN training, the state-of-the-art methods are mostly based on the WGAN framework \cite{arjovsky2017wasserstein}, which has the following objective function:
\begin{equation} \label{eq:wgan}
% \begin{split}
    \min_G \max_D~~\mathbb{E}_{x \sim P_{r}}[D(x)] 
     - \mathbb{E}_{\tilde{x} \sim P_{g}}[D(\tilde{x})],
% \end{split}
\end{equation}
where $x$ represents real data with its distribution $P_{r}$ and $\tilde{x} = G(z)$ represents generated data from the distribution $P_g$ such that $z \sim p(z)$s. \ourmodel is also trained by optimizing Eq. (\ref{eq:wgan}), not on real data samples ($x \sim P_{r}$) but on parameters sampled from the task-specific BNN (\textit{i.e.,} $\theta_\B \sim \mathcal{N}(\mu_{\theta_\B}, \Sigma_{\theta_\B})$) of the first task. In addition, rather than directly feeding entire models of $\theta_\B$ to \ourmodel, we split $\theta_\B$ into a set of equal-sized chunks $\Phi_{\B}=\left\{ \phi^{1}_\B,\cdot\cdot\cdot,\phi^{m}_\B \right\}$ s.t. $m = \lceil|\theta_\B|/|\phi^{i}_\B|\rceil$, as used in a few studies \cite{HaDL17,JosephB20,OswaldHSG20}. Then, these sampled parameter chunks, each conditioned on its chunk-id, are fed into \ourmodel as training samples. This chunking technique reduces the output dimensionality of the meta-model to be the size of each chunk (\textit{e.g.,} $|\phi^{i}_\B|=2000$) from the entire model size.

\smalltitle{Incremental Learning on \ourmodel} 
Then, for the $t$-th task s.t. $t > 1$, the previous generator $G^{t-1}$ is used to generate synthetic parameters of the previous tasks from $1$ to $t-1$, and then $G^t$ is jointly trained with the combined set of current and generated chunks. When \ourmodel learns the distribution of parameter chunks $\Phi^t_\B \sim \mathcal{N}(\mu_{\Phi_{\B}^t}, \Sigma_{\Phi_{\B}^t})$ sampled from the BNN for task $t$, \textit{real} chunks $\Phi^{1:t-1}_\B$ from the previously learned BNNs are replaced by generated chunks $\Phi^{1:t-1}_\G=\{G^{t-1}(z,c)\}$, where $z$ is a latent vector sampled from $\mathcal{N}(0,1)$ and $c$ is a chunk-id randomly chosen from all chunk-ids of the previous tasks. As shown in Figure \ref{fig:main_illust}, the whole training dataset $\Phi^{1:t}_\R$, which corresponds to $x \sim P_r$ in Eq. (\ref{eq:wgan}), is the union of sampled chunks $\Phi^{t}_\B$ and generated chunks $\Phi^{1:t-1}_\G$, that is, $\Phi^{1:t}_\R = \Phi^{t}_\B \cup \Phi^{1:t-1}_\G$. Thus, \ourmodel at the $t$-th task is trained to generate $\Phi^{1:t}_\G$ that are the chunks for all tasks learned so far with the following objective function:
\begin{equation} \label{eq:gamm_loss}
\begin{split}
    \min_G \max_D~~\mathbb{E}_{\Phi^{1:t}_\R \sim P_r}[D(\Phi^{1:t}_\R)] 
    - \mathbb{E}_{\Phi^{1:t}_\G \sim P_g}[D(\Phi^{1:t}_\G)].
\end{split}
\end{equation}

In the end of the knowledge consolidation step, $G^t$ is only retained in \ourmodel for making inference as well as learning the next task, and all the other components are not stored.

\subsection{Knowledge Recall} \label{subsec:recall}

When the human brain recalls some information previously remembered, the corresponding knowledge is somehow generated from long-term memory \cite{schacter2007cognitive}, and brought into working memory for use. Similarly, whenever making inference for each task in our framework, the required knowledge, that is, the task-specific model itself, can be generated by \ourmodel. More precisely, given a task-identifier $k \in [1,T]$, \ourmodel produces a set of parameter chunks $\Phi^{\,k}_\G = \{G(z, c^k)\}$, each of which is conditionally generated by a chunk-id $c^k$ for task $k$. The generated chunks are then assembled into the corresponding task-specific model $\theta^{\,k}_\G$, as illustrated in Figure \ref{fig:main_illust}, which can be immediately used to make inference. This generated model is expected to output similar results to those of its original task-specific BNN, that is, $p(y|x,\theta_\G^{\,k}) \approx p(y|x,\theta_\B^{\,k})$.

\smalltitle{Task-Agnostic Inference} 
Although this paper focuses on the above TIL scenario using task-identifiers, the generation scheme of \ourmodel is also applicable to make prediction without task-identifiers at inference time. To this end, we can choose a model with the highest confidence out of task-specific models over all the tasks previously learned, and then use its output as our final prediction. We adopt the strategy of selecting a model with the smallest entropy $\mathcal{H}$ \cite{HenningCDOTEKGS21}, indicating the highest confidence, and finally return $p(y|x,\theta_\G^{\hat{k}})$ s.t. $\hat{k} = \argmin\limits_{j \in [1,T]}\mathcal{H}\left[p(y|x, \theta^{\,j}_\G)\right]$.

%--------------------------------------------------------------------------------------

\begin{table*}
\centering
\renewcommand{\arraystretch}{1.15}
\resizebox{\textwidth}{!}{%
\begin{tabular}{llccccrcccccr}
\Xhline{2.5\arrayrulewidth}
 & & \multicolumn{5}{c}{\textbf{Split CIFAR-10}} &  & \multicolumn{5}{c}{\textbf{Split CIFAR-100}}  \\ \cline{3-7} \cline{9-13} 
 & & \multicolumn{3}{c}{Task-aware} & Task-agnostic &  & & \multicolumn{3}{c}{Task-aware} & Task-agnostic &  \\ \cline{3-7} \cline{9-13}
\multicolumn{1}{c}{Method} & \multicolumn{1}{c}{Strategy} & ACC & BWT & LA & ACC & \multicolumn{1}{c}{Memory} &  & ACC & BWT & LA & ACC & \multicolumn{1}{c}{Memory} \\ \hline
A-GEM    & Mem (I)     &  85.97 &  -11.96  &  95.54 &  20.15 &  $\approx$ 1.2 M &  &  56.36 &         -35.15 & \textbf{87.99} &   9.80 & $\approx$ 18.9 M \\
ER       & Mem (I)     &  88.57 &   -6.22  &  93.55 &  50.32 &  $\approx$ 1.2 M &  &  80.85 &          -5.52 &          85.82 &  31.82 & $\approx$ 18.9 M \\
DGR      & Gen \,\ (I) &  70.38 &  -17.31  &  84.23 &  13.83 &       14.3 M &  &  16.55 &         -48.12 &          59.86 &  12.44 &       25.0 M \\
BI-R     & Gen \,\ (F) &  94.99 &   -0.18  &  95.14 &  55.19 &       13.3 M &  &  82.33 &          -0.64 &          82.90 &  22.83 &       13.3 M \\
HNET & Gen \,\ (P) &  94.53 &   -0.45  &  94.89 &  56.67 &        1.0 M &  &  86.50 &          -0.26 &          86.73 &  41.63 &        5.0 M \\
PR-BBB   & Gen \,\ (P) &  95.30 &   -0.28  &  95.52 &  58.94 &       20.2 M &  &  84.78 & \textbf{-0.05} &          84.82 &  41.66 &       14.7 M \\ \hline
\textbf{\ourmodel} & Gen \,\ (P) & \textbf{96.73} & \textbf{-0.13} & \textbf{96.84} & \textbf{78.49} & 0.9 M & & \textbf{87.19} & \underline{-0.22} & \underline{87.39} & \textbf{44.75} &  5.1 M \\
\Xhline{2.5\arrayrulewidth}
\end{tabular}%
}
\caption{Performance of replay-based methods, where we report the mean of ACC, BWT, and LA over five runs with different task orders. The approximation symbol ($\approx$) represents the memory usage considering both parameters and exemplars.}
\label{tab:exp_memory}
\end{table*}
 % (standard deviations are reported in the Appendix)

\begin{table*}
\centering
\scriptsize
\renewcommand{\arraystretch}{1.15}
\resizebox{0.95\textwidth}{!}{%
\begin{tabular}{lcccccccc}
\Xhline{2.5\arrayrulewidth}
& \multicolumn{2}{c}{\textbf{Permuted MNIST}} &  & \multicolumn{2}{c}{\textbf{Split CIFAR-100}} &  & \multicolumn{2}{c}{\textbf{5-Datasets}} \\ \cline{2-3} \cline{5-6} \cline{8-9} 
Methods & ACC & BWT &  & ACC & BWT &  & ACC & BWT \\ \hline
EWC   & 92.01 ± 0.56$^\dagger$ & \textbf{-0.03} ± 0.00$^\dagger$ &  & 71.17 ± 0.46 & -2.23 ± 0.47 &  & 88.64 ± 0.26$^\dagger$ & -0.04 ± 0.01$^\dagger$ \\
HAT   & - & - &  & 73.25 ± 0.30 & -2.79 ± 0.43 &  & 91.32 ± 0.18$^\dagger$ & \textbf{-0.01} ± 0.00$^\dagger$ \\
A-GEM & 83.56 ± 0.16$^\dagger$ & -0.14 ± 0.00$^\dagger$ &  & 63.98 ± 1.22$^\dagger$ & -0.15 ± 0.02$^\dagger$ &  & 84.04 ± 0.33$^\dagger$ & -0.12 ± 0.01$^\dagger$ \\
ER    & 87.24 ± 0.53$^\dagger$ & -0.11 ± 0.01$^\dagger$ &  & 71.73 ± 0.63$^\dagger$ & \textbf{-0.06} ± 0.01$^\dagger$ &  & 88.31 ± 0.22$^\dagger$ & -0.04 ± 0.00$^\dagger$ \\
GPM   & 93.91 ± 0.16$^\dagger$ & \textbf{-0.03} ± 0.00$^\dagger$ &  & 72.45 ± 0.54 & -0.43 ± 0.43 &  & 91.22 ± 0.20$^\dagger$ & \textbf{-0.01} ± 0.00$^\dagger$ \\ \hline
\textbf{\ourmodel}  & \textbf{96.40} ± 0.02 & -0.68  ± 0.12 &  & \textbf{74.90} ± 0.94 & -0.18 ± 0.08 &  & \textbf{93.68} ± 0.20 & -0.31 ± 0.14  \\
\Xhline{2.5\arrayrulewidth}
\end{tabular}%
}
\caption{Performance of the proposed framework and baseliness. Mean and standard deviation of ACC and BWT over five runs with different task orders are reported. The values with $\dagger$ are from \cite{SahaG021}.}
\label{tab:baselines}
\end{table*}

\section{Experiments} \label{sec:experiment}

This section evaluates our framework by comparing with the existing replay-based methods as well as the other CL baselines from the perspective of the stability-plasticity dilemma.

\subsection{Experimental Settings for CL Benchmarks}

\smalltitle{Performance Metrics}
We evaluate CL performance using three metrics: \textit{average accuracy} (ACC), \textit{backward transfer} (BWT) and \textit{learning accuracy} (LA) as follows: $ACC=\frac{1}{T}\sum_{i=1}^T R_{T,i}, \; BWT=\frac{1}{T-1}\sum_{i=1}^{T-1} R_{T,i}-R_{i,i}, \; LA=\frac{1}{T}\sum_{i=1}^{T} R_{i,i},$
where $T$ is the total number of tasks and $R_{i,j}$ is the classification accuracy on the $j$-th task after learning the $i$-th task. ACC is the average accuracy of all the tasks in the end. BWT measures how much we forget the past knowledge, where the more negative values, the more forgetting occurs. LA is to measure the model's plasticity \cite{RiemerCALRTT19}, by taking the average accuracy for each task right after it has been learned. Our final goal is to achieve highest ACC with near-zero BWT yet large LA values.

\smalltitle{Datasets}
We consider the datasets commonly used in CL benchmarks, namely Split CIFAR-10 and Split CIFAR-100 for comparison with replay-based methods, and PMNIST, Split CIFAR-100 and 5-Datasets \cite{ebrahimi2020adversarial} for comparison with the other CL baselines. Split CIFAR-10 is constructed by dividing the original CIFAR-10 dataset into 5 tasks, each with 2 classes. Similarly, Split CIFAR-100 consists of 10 tasks, each with 10 classes.
PMNIST is a variant of MNIST, where each task has a different random permutation on image pixels. 5-Datasets consists of five different datasets: CIFAR-10, MNIST, SVHN, FMNIST \cite{xiao2017fashion}, and notMNIST \cite{bulatov2011notmnist}, each of which is learned as a single task.

\smalltitle{Baselines}
We test various replay-based methods, which are classified into two categories: memory replay (Mem) and generative replay (Gen). Also, their replaying targets can be in one of the following three levels: input-level (I), feature-level (F), and parameter-level (P). Both A-GEM \cite{ChaudhryRRE19} and ER \cite{chaudhryER_2019} replay input-level representations with a fairly large memory buffer of 100 exemplars per class \cite{van2022three}. DGR \cite{ShinLKK17} and BI-R \cite{van2020brain} belong to generative replay with input-level and feature-level representations, respectively. HNET \cite{OswaldHSG20} and PR-BBB \cite{HenningCDOTEKGS21} replay parameter-level representations via hypernetworks, which are the state-of-the-arts in replay-based methods. We consider the other types of baselines; regularization-based EWC \cite{kirkpatrick2017overcoming}, architecture-based HAT \cite{SerraSMK18}, and memory-based GPM \cite{SahaG021}.

\smalltitle{Model Architectures}
We use a simple GAN architecture for \ourmodel, where the generator $G$ consists of 100 input units followed by two fully-connected (FC) layers of 200 units and the discriminator $D$ has one FC layer of 256 units followed by a binary output head. The output dimensionality of $G$ varies depending on the chunk size. For inference models to make prediction, \ourmodel generates lightweight neural networks, which are reduced versions of the original architectures (e.g., 0.15 times the original size of ResNet-32) that have been commonly used in the literature. Those original architectures include: (1) ResNet-32 for Split CIFAR-10 \cite{HenningCDOTEKGS21}, (2) ResNet-18 \cite{HenningCDOTEKGS21} and 5-layer AlexNet \cite{SahaG021} for Split CIFAR-100, (3) a FC network with two layers of 100 units \cite{SahaG021} for PMNIST, and (4) reduced ResNet-18 \cite{SahaG021} for 5-Datasets. Some baselines utilize their own architectures, such as BI-R \cite{van2020brain} with an autoencoder-based model, and HNET \cite{OswaldHSG20} and PR-BBB \cite{HenningCDOTEKGS21} with hypernetwork-based meta-models \cite{HenningCDOTEKGS21}. More experimental details are presented in the Appendix.

\subsection{CL Performance Comparison and Discussion}

\smalltitle{Comparison with Replay-Based Methods}
We compare the performance of replay-based methods in both task-aware and task-agnostic settings. Table \ref{tab:exp_memory} summarizes the experimental results. Our framework with \ourmodel outperforms all the baselines, consistently showing the highest ACC that indicates the best stability-plasticity balance. In task-aware settings, \ourmodel achieves almost zero-forgetting with its near-zero BWT values, while it also shows the excellent plasticity with high LA values. The performance gap between \ourmodel and the compared methods becomes larger in task-agnostic scenarios, which shows the highest effectiveness of \ourmodel even without task-identifiers at inference time. As our expectation in analysis on representation complexity, the methods of replaying either exemplars (A-GEM and ER) or pseudo-samples (DGR) are observed to suffer from severe forgetting, showing most negative BWT values. This confirms that replaying input-level representations is indeed not effective at knowledge preservation due to their high complexity. BI-R using feature-level replay has higher stability with better BWT values, but shows limited LA values due to the stability-plasticity dilemma that is partially unresolved by replaying feature vectors. HNET and PR-BBB also exhibit better BWT compared to input-level replay methods, but their plasticity is worse than \ourmodel in terms of LA. This is likely due to their one-phase training scheme within a hypernetwork, where the process of learning new knowledge can be hindered by the process of retaining past knowledge. This one-phase training seems to make worse effect on the performance in task-agnostic settings. In contrast, \ourmodel manages to achieve the best performance not only in task-aware scenarios but also in task-agnostic scenarios by separating the mechanisms of learning new knowledge from preserving old knowledge.

\smalltitle{Memory Usage}
Table \ref{tab:exp_memory} also shows the amount of memory required for each replay-based model to maintain the learned knowledge. The memory usage is measured by the total number of parameters (or values) in the main model, the generative model, or the stored exemplars. \ourmodel shows better efficiency in memory usage, compared to the memory replay and generative replay methods, and uses a similar amount of memory to what HNET takes, while ensuring always the best performance.

\smalltitle{Comparison with CL Baselines}
We compare our framework with the other baselines with respect to ACC and BWT. Table \ref{tab:baselines} summarizes the experimental results, where PMNIST is a task-agnostic setting and the others follows a task-aware setting. Our framework outperforms all the baselines with clear margins in ACC. Most methods achieve BWT close to 0 while their ACC is not as good as that of our framework, which implies that their techniques to alleviate forgetting have impaired plasticity of models. In contrast, our method achieves the highest ACC thanks to its high plasticity via the separation of working and long-term memory.

%--------------------------------------------------------------------------------------

\section{Conclusion} \label{sec:conclusion}
In this paper, we proposed a novel framework to resolve the stability-plasticity dilemma in continual learning, where the task-specific knowledge is separately learned, and then consolidated into the meta-model. Based on our analysis on representation complexity, we found that the parameter-level representation is a proper form of knowledge to be maintained, and thereby proposing \ourmodel that can re-generate task-specific models themselves to make prediction at inference time. Our experimental results confirm that the proposed framework is capable of achieving the best balance between stability and plasticity without a large amount of memory consumption, outperforming the existing methods in both task-aware and task-agnostic scenarios.

%--------------------------------------------------------------------------------------

\section{Acknowledgments}
This work was supported in part by Institute of Information \& communications Technology Planning \& Evaluation (IITP) grants funded by the Korea government(MSIT) (No.2022-0-00448, Deep Total Recall: Continual Learning for Human-Like Recall of Artificial Neural Networks, No.RS-2022-00155915, Artificial Intelligence Convergence Innovation Human Resources Development (Inha University)), in part by the National Research Foundation of Korea (NRF) grants funded by the Korea government (MSIT) (No.2021R1F1A1060160, No.2022R1A4A3029480), and in part by INHA UNIVERSITY Research Grant.

%--------------------------------------------------------------------------------------

\bibliography{aaai24}

\input{appendix}

\end{document}

%% file: appendix.tex
% \newpage
\onecolumn
\setcounter{table}{0}
\setcounter{figure}{0}
\renewcommand{\thetable}{A\arabic{table}}  
\renewcommand{\thefigure}{A\arabic{figure}}
\appendix
\section*{Appendix: Recall-Oriented Continual Learning with Generative Adversarial Meta-Model} \label{sec:appendix}

In this appendix, we present (1) detailed processes of knowledge acquisition and consolidation procedures, respectively, and (2) additional experimental details of representation complexity analysis and performance comparison using CL benchmarks.

\subsection{The Process of Knowledge Acquisition}
\begin{algorithm}[h]
    \caption{Knowledge acquisition with SWAG at the $t$-th task } \label{alg:cnstrt}
    \begin{algorithmic}[1]
    \REQUIRE initial BNN parameters  $\theta_\B^{\,0}$; training data $\mathcal{D}^t$; starting epoch of computing moments $n_{s}$; number of epochs $N_{epoch}$
        \STATE $\overline{\theta_\B} \gets \theta_\B^{\,0}, \; \overline{{\theta_\B}^2} \gets {\theta_\B^{\,0}}^2, \; n \gets 0$
        \FOR{epochs $i=1,\cdot\cdot\cdot,N_{epoch}$}
            \STATE $\theta_\B^{\;i} \gets SGD(\nabla\mathcal{L}_{ce}(\theta_\B, \mathcal{D}^t))$
            \IF{$i \ge n_{s}$}
                \STATE $\overline{\theta_\B} \gets \frac{n\overline{\theta_\B}+\theta_\B^i}{n+1}, \ \overline{{\theta_\B}^2} \gets \frac{n\overline{{\theta_\B}^2}+{\theta_\B^i}^2}{n+1}, \ n \gets n+1$ \alglinelabel{line:runavg}
            \ENDIF
            \STATE $\mu_{\theta_{\B}^{t}} \gets \overline{\theta_\B}, \ \Sigma_{\theta_\B^{t}} \gets diag(\overline{{\theta_\B}^2} - {\overline{\theta_\B}}^{2})$
        \ENDFOR
        \STATE \textbf{return} $\left\{ \mu_{\theta_{\B}^{t}}, \Sigma_{\theta_\B^{t}} \right\}$
    \end{algorithmic}
\end{algorithm}
The process of knowledge acquisition with stochastic weight averaging Gaussian (SWAG) \cite{IzmailovPGVW18,MaddoxIGVW19} is presented in Algorithm \ref{alg:cnstrt}.
SWAG is based on the assumption that there is useful information in the history of stochastic gradient descent (SGD) so that we can capture a Gaussian posterior distribution of model parameters. The loss for training SWAG is negative log-likelihood that is a cross-entropy loss  $\L_{ce}=-\sum_{x,y\in\D^{t}} y\log{p(x)}$ in the classification. During SGD update, the first and second moments of parameters are estimated with parameter values over the last $N$ epochs as $~\loline{\theta_\B}\ = \frac{1}{N}\sum_{i=1}^N \theta^{\,i}_\B$ and $~\soline{{\theta_\B}^2} = \frac{1}{N}\sum_{i=1}^N {\theta^{\,i}_\B}^2$, respectively, where $\theta_\B^{\; i}$ is the set of parameter values of the $i$-th epoch and ${\theta^{\,i}_\B}^2$ is the set of their element-wise squares. Using these first and second moments, the mean and covariance of the Gaussian posterior distribution are estimated as $\mu_{\theta_\B^{t}} = \ \loline{\theta_\B}~$ and $\Sigma_{\theta_\B^{t}} = diag(\ \soline{{\ \theta_\B}^2}\, - \, {\loline{\theta_\B}}^{\,2})$. In practice, the $\ \loline{\theta_\B}\ $ and $\ \soline{{\theta_\B}^{2}}\ $ are computed by running average as in Line \ref{line:runavg} of Algorithm \ref{alg:cnstrt}. Consequently, the posterior distribution of the BNN $\theta_{\B}^{\,t}$ is approximated by the Gaussian distribution $\mathcal{N}(\mu_{\theta_\B^t}, \Sigma_{\theta_\B^t})$ from which we can sample the task-specific parameters for task $t$.

\subsection{The Process of Knowledge Consolidation}
\begin{algorithm}[h]
    \caption{Knowledge consolidation on \ourmodel at the $t$-th task} \label{alg:consol}
    \begin{algorithmic}[1]
        \REQUIRE lightweight BNN $\theta_\B^{\,t}$; \ourmodel parameterized by $\theta_\G$; previous generator $G^{t-1}$; chunk size $m$;
        \STATE Randomly initialize $\theta_\G$ \alglinelabel{line:init}
        \WHILE{$\theta_\G$ has not converged}
            \STATE Sample parameters $\theta_\B^{\,t} \sim \mathcal{N}(\mu_{\theta_\B^{t}}, \Sigma_{\theta_\B^{t}})$ for task $t$, latent vector $z \sim \mathcal{N}(0,1)$, random chunk-ids $c$ for previous tasks
            \STATE $\Phi^t_\B \gets chunking(\theta_\B^{\,t})$
            \IF{$t=1$}
                \STATE $\Phi^1_\R \gets \Phi_\B^t$
            \ELSE
                \STATE $\Phi^{1:t-1}_\G \gets \{G^{t-1}(z,c)\}$, for random chunk-id $c$ for previous tasks s.t. $c \in [1, (t-1)m]$
                \STATE $\Phi^{1:t}_{\R} \gets \Phi_{\B}^t \cup \Phi_{\G}^{1:t-1}$
            \ENDIF
            \STATE Train $\theta_\G$ with $\Phi^{1:t}_{\R}$ via generative adversarial learning with Eq (\ref{eq:D_loss}) \& Eq (\ref{eq:G_loss})
        \ENDWHILE
    \end{algorithmic}
\end{algorithm}
The detailed process of training \ourmodel at task $t$ is summarized in Algorithm \ref{alg:consol}. As shown in Line \ref{line:init} of Algorithm \ref{alg:consol}, all parameters of \ourmodel are randomly initialized before training. We do not train \ourmodel with all sampled parameters stored, but with parameters newly sampled from $\theta_\B^{\,t}$ at every training iteration. During the training loop, generative replay on model parameters is performed as described in our methodology, and the weights of \ourmodel $\theta_\G$ is trained through generative adversarial learning \cite{NIPS2014_5ca3e9b1, GulrajaniAADC17} by minimizing the loss functions in Eq. (\ref{eq:D_loss}) and Eq. (\ref{eq:G_loss}).

In generative adversarial learning, the discriminator $D$ and generator $G$ are trained alternately. When training discriminator $D$, we use gradient penalty $\L_{gp} =\mathbb{E}_{\hat{\Phi}}[(\lVert \nabla_{\hat{\Phi}}D(\hat{\Phi}) \rVert-1)^2]$ that enforces Lipschitz constraint on $D$ as done in WGAN-GP \cite{GulrajaniAADC17}, where $\hat{\Phi}$ is an interpolation between $\Phi_\R^{1:t}$ and $\Phi_\G^{1:t}$, which is defined by $\hat{\Phi}=\epsilon\, \Phi_\R^{1:t} + (1-\epsilon)\Phi_\G^{1:t}$. Additionally, when training generator $G$, replay alignment loss $\L_{ra} = \mathbb{E}_{z \sim \mathcal{N}(0, 1)} [\lVert G^{t}(z,c) - G^{t-1}(z,c) \lVert^{2}_{2}]$ is added, which minimizes the Euclidean distance between the outputs of $G^{t}$ and $G^{t-1}$, as used in MeRGAN-RA \cite{WuHLWWR18}. 

\smalltitle{Loss Functions of $D$ and $G$}
The loss functions with respect to $D$ and $G$ are described in Eq (\ref{eq:D_loss}) and Eq (\ref{eq:G_loss}), respectively, where $\lambda_{gp}$ and $\lambda_{ra}$ are hyperparameters.
\begin{equation} \label{eq:D_loss}
    \L_D = - \mathbb{E}_{\Phi_\R^{1:t} \sim P_r}[D(\Phi_\R^{1:t})] + \mathbb{E}_{\Phi_\G^{1:t} \sim P_g}[D(\Phi_\G^{1:t})] + \lambda_{gp}\, \L_{gp},
\end{equation}
\begin{equation} \label{eq:G_loss}
    \L_G = - \mathbb{E}_{\Phi_\G^{1:t} \sim P_g}[D(\Phi_\G^{1:t})] + \lambda_{ra}\, \L_{ra},
\end{equation}

\subsection{Additional Experimental Details and Results}

\begin{table*}[h]
\centering
\renewcommand{\arraystretch}{1.2}
\resizebox{0.8\textwidth}{!}{%
\begin{tabular}{lrrrrrrrrrrr}
\Xhline{3\arrayrulewidth}
& \multicolumn{3}{c}{\textbf{MNIST}} & \multicolumn{1}{l}{} & \multicolumn{3}{c}{\textbf{SVHN}} & \multicolumn{1}{l}{} & \multicolumn{3}{c}{\textbf{CIFAR-10}} \\ \cline{2-4} \cline{6-8} \cline{10-12} 
 & \multicolumn{1}{c}{Input} & \multicolumn{1}{c}{Feature} & \multicolumn{1}{c}{Param.} & \multicolumn{1}{c}{} & \multicolumn{1}{c}{Input} & \multicolumn{1}{c}{Feature} & \multicolumn{1}{c}{Param.} & \multicolumn{1}{c}{} & \multicolumn{1}{c}{Input} & \multicolumn{1}{c}{Feature} & \multicolumn{1}{c}{Param.} \\ \hline
Dim.  &    784   &       32   &      7000  & &      3072     &     32   &     7000   & &      3072    &       32  &    7000   \\
$\S$  &  0.20    &     0.87   &     0.94   & &     0.01      &    0.70  &     0.91   & &      0.04    &     0.56  &    0.90  \\
$\V$  & 0.14 M   &  2075.91   &  2868.37   & &   5371.50 M   & 1589.30  &  2871.64   & &  11748.04 M  &  1166.91  & 2871.66  \\
% $\C$  & 0.11 M   &   266.49   &    160.63  & &   5344.64 M   &  480.73  &   246.96   & &  11278.12 M  &   515.66  &  295.78  \\
\Xhline{3\arrayrulewidth}
\end{tabular}%
}
\caption{Separability ($\S$) and Volume ($\V$) of input images, feature vectors, and parameter chunks on MNIST, SVHN and CIFAR-10, where the number of groups is equally 10 and ResNet-32 is used for all datasets.}
\label{tab:complexity}
\end{table*}

\begin{figure*}[h]
\centering
\subfigure[\label{fig:mnist_input}MNIST inputs]{\includegraphics[width=0.24\columnwidth]{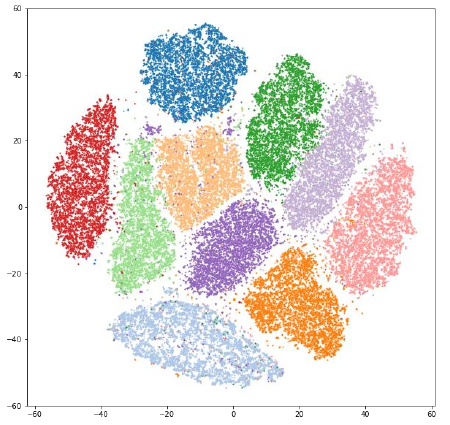}}
\subfigure[\label{fig:mnist_param}MNIST parameters]{\includegraphics[width=0.24\columnwidth]{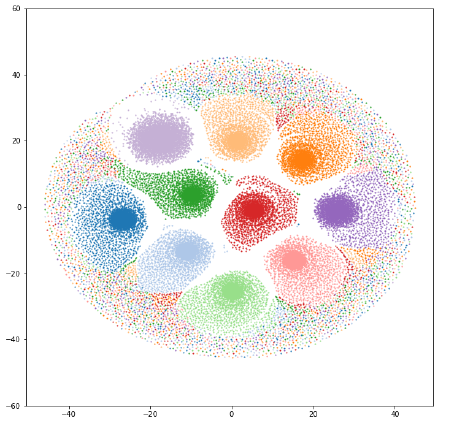}}
\subfigure[\label{fig:cifar10_input}CIFAR-10 inputs]{\includegraphics[width=0.24\columnwidth]{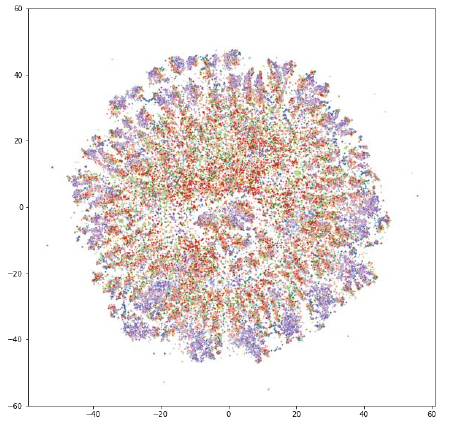}}
\subfigure[\label{fig:cifar10_param}CIFAR-10 parameters]{\includegraphics[width=0.24\columnwidth]{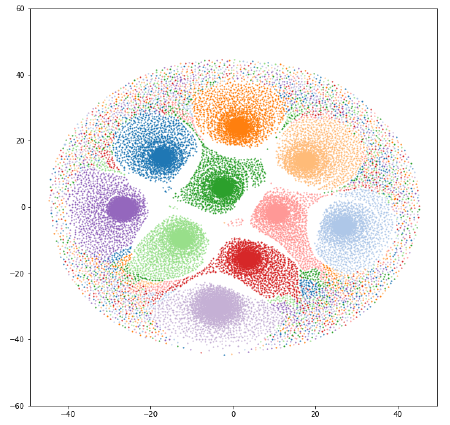}}
\caption{T-SNE visualizations of input images and their corresponding parameter chunks, where each point, indicating either an image or a chunk, has a different color for a class or a chunk-id.}\label{fig:tsne_visual}
\end{figure*}

\subsubsection{Comparison on Representational Complexity} \label{sec:experiment:complexity}

Our metrics on representational complexity can measure how complex each representation type is to learn the knowledge behind a particular dataset. Table \ref{tab:complexity} presents the volume, separability and dimensionality of three representation types. As input-level representations reflect datasets themselves, we can first see that CIFAR-10 is more complex than SVHN, and MNIST is the simplest in terms of volume ($\V$) as well as separability ($\S$). In particular, 
the volume of MNIST is notably smaller than those of the other two datasets, as CIFAR-10 and SVHN have more diverse images than MNIST. This implies that a dataset with more diverse samples will make memory replay more challenging. Feature vectors provide a concise representation of knowledge, with significantly lower complexity compared to images. However, they may not be the best choice for replaying, as the front part of a model cannot be updated using feature vectors, leaving the stability-plasticity dilemma unresolved. Parameter chunks exhibit the outstanding separability and thus show a lower complexity even than that of feature vectors. This is a notable result in that parameter chunks are distributed in a higher-dimensional space than feature vectors. Overall, we can conclude that learned parameters are better than not only images but also feature vectors in terms of our complexity metric, and this trend is consistently observed in the CL performance comparison.

\subsubsection{Visualization of Input-Level and Parameter-Level Representations}
Figure \ref{fig:tsne_visual} visualizes how input images and their corresponding parameter chunks are distributed in their latent space, where each color represents a different class or chunk-id. It is observed that MNIST has well-separated data points within their class, while the images in CIFAR-10 are irregularly scattered without any clear distinctions among them. In contrast, parameter chunks in both MNIST and CIFAR-10 are well-separated in their latent spaces.

\begin{table}[h!]
\centering
\renewcommand{\arraystretch}{1.15}
\resizebox{0.85\textwidth}{!}{%
\begin{tabular}{ccclccc}
\Xhline{2.0\arrayrulewidth}
\multicolumn{1}{l}{} & \multicolumn{2}{c}{\textbf{Comparison with Replay-Based Methods}} &  & \multicolumn{3}{c}{\textbf{Comparison with Baselines}} \\ \cline{2-3} \cline{5-7} 
 & Split CIFAR-10 & Split CIFAR-100 &  & PMNIST & Split CIFAR-100 & 5 Datasets \\ \cline{1-7} 
Original model   &  ResNet-32    &  ResNet-18   &  &    FC Network  &   5-layer Alexnet &   Modified ResNet-18   \\
Compression rate &  0.15         &  0.04        &  &    1.00        &   0.06            &   0.06 \\
Epoch            &  120          &  250         &  &    5           &   200             &   100 \\
(Initial) LR     &  0.1          &  0.1         &  &    0.01        &   0.1             &   0.1 \\
\ourmodel LR     &  0.0002       &  0.005       &  &    0.001       &   0.0002          &   0.0002 \\
Chunk size       &  2000         &  25000       &  &    10000       &   25000           &   5000 \\
$\lambda_{gp}$   &  10.0         &  10.0        &  &    10.0        &   10.0            &   10.0 \\
$\lambda_{ra}$   &  5.0          &  15.0        &  &    50.0        &   10.0            &   30.0 \\
\Xhline{2.0\arrayrulewidth}
\end{tabular}%
}
\caption{Experimental details including the hyperparameters of \ourmodel}
\label{tab:exp_detail}
\end{table}

\begin{table}[h!]
\centering
\renewcommand{\arraystretch}{1.15}
\resizebox{\textwidth}{!}{%
\begin{tabular}{llccccrcccccr}
\Xhline{2.5\arrayrulewidth}
 & & \multicolumn{5}{c}{\textbf{Split CIFAR-10}} &  & \multicolumn{5}{c}{\textbf{Split CIFAR-100}}  \\ \cline{3-7} \cline{9-13} 
 & & \multicolumn{3}{c}{Task-aware} & Task-agnostic &  & & \multicolumn{3}{c}{Task-aware} & Task-agnostic &  \\ \cline{3-7} \cline{9-13}
\multicolumn{1}{c}{Method} & \multicolumn{1}{c}{Strategy} & ACC & BWT & LA & ACC & \multicolumn{1}{c}{Memory} &  & ACC & BWT & LA & ACC & \multicolumn{1}{c}{Memory} \\ \hline
A-GEM    & Mem (I)     &  85.97 $\pm$ 1.80 &  -11.96 $\pm$ 1.29 &  95.54 $\pm$ 1.24 &  20.15 $\pm$ 0.64 &  $\approx$ 1.2 M &  &  56.36 $\pm$ 0.94 &         -35.15 $\pm$ 1.02 & \textbf{87.99} $\pm$ 0.46 &   9.80 $\pm$ 0.24 & $\approx$ 18.9 M \\
ER       & Mem (I)     &  88.57 $\pm$ 4.09 &   -6.22 $\pm$ 1.99 &  93.55 $\pm$ 3.38 &  50.32 $\pm$ 3.70 &  $\approx$ 1.2 M &  &  80.85 $\pm$ 0.63 &          -5.52 $\pm$ 0.35 &          85.82 $\pm$ 0.64 &  31.82 $\pm$ 0.99 & $\approx$ 18.9 M \\
DGR      & Gen \,\ (I) &  70.38 $\pm$ 3.09 &  -17.31 $\pm$ 3.67 &  84.23 $\pm$ 1.16 &  13.83 $\pm$ 2.13 &  14.3 M          &  &  16.55 $\pm$ 1.80 &         -48.12 $\pm$ 3.81 &          59.86 $\pm$ 4.06 &  12.44 $\pm$ 0.90 &       25.0 M \\
BI-R     & Gen \,\ (F) &  94.99 $\pm$ 0.13 &   -0.18 $\pm$ 0.11 &  95.14 $\pm$ 0.10 &  55.19 $\pm$ 3.62 &  13.3 M          &  &  82.33 $\pm$ 0.06 &          -0.64 $\pm$ 0.29 &          82.90 $\pm$ 0.24 &  22.83 $\pm$ 1.83 &       13.3 M \\
HNET & Meta (P)    &  94.53 $\pm$ 0.67 &   -0.45 $\pm$ 0.38 &  94.89 $\pm$ 0.59 &  56.67 $\pm$ 3.01 &  1.0 M           &  &  86.50 $\pm$ 0.53 &          -0.26 $\pm$ 0.16 &          86.73 $\pm$ 0.61 &  41.63 $\pm$ 0.65 &        5.0 M \\
PR-BBB   & Meta (P)    &  95.30 $\pm$ 0.26 &   -0.28 $\pm$ 0.17 &  95.52 $\pm$ 0.12 &  58.94 $\pm$ 2.20 &  20.2 M          &  &  84.78 $\pm$ 0.35 & \textbf{-0.05} $\pm$ 0.12 &          84.82 $\pm$ 0.34 &  41.66 $\pm$ 0.38 &       14.7 M \\ \hline
\textbf{\ourmodel} & Meta (P) & \textbf{96.73} $\pm$ 0.08& \textbf{-0.13} $\pm$ 0.04 & \textbf{96.84} $\pm$ 0.07 & \textbf{78.49} $\pm$ 0.82 & 0.9 M & & \textbf{87.19} $\pm$ 0.27 & \underline{-0.22} $\pm$ 0.25 & \underline{87.39} $\pm$ 0.21 & \textbf{44.75} $\pm$ 1.46 &  5.1 M \\
\Xhline{2.5\arrayrulewidth}
\end{tabular}%
}
\caption{Performance of the our framework and replay-based methods on CL benchmarks. Replay strategies for each method are reported. We report the mean of ACC, BWT, and LA with the standard deviation over five runs with different task orders. The approximation symbol represents the estimated memory size that takes into account the size of the memory buffer.}
\label{tab:app:exp_memory}
\end{table}

\begin{table}[h!]
\centering
\renewcommand{\arraystretch}{1.15}
\resizebox{\textwidth}{!}{%
\begin{tabular}{lcccrrrrr}
\Xhline{2.0\arrayrulewidth}
& \multicolumn{1}{l}{} & \multicolumn{1}{l}{} & \multicolumn{1}{l}{} & \multicolumn{5}{c}{\textbf{Memory usage on Split   CIFAR-10}} \\ \cline{5-9} 
\multicolumn{1}{c}{Method} & Strategy  & Replay Target  &  & \multicolumn{1}{c}{Main model} & \multicolumn{1}{c}{Memory buffer} & \multicolumn{1}{c}{Generative model} & \multicolumn{1}{c}{Meta-model} & \multicolumn{1}{c}{Total} \\ \hline
A-GEM      & Memory replay     & Input      & &   0.5 M  &    $\approx$ 0.8 M  &            &          &  $\approx$ 1.2 M \\
ER         & Memory replay     & Input      & &   0.5 M  &    $\approx$ 0.8 M  &            &          &  $\approx$ 1.2 M \\
DGR        & Generative replay & Input      & &   0.5 M  &                     &   13.8 M   &          &           14.3 M \\
BI-R       & Generative replay & Feature    & &          &                     &   13.3 M   &          &           13.3 M \\
HNET   & Generative replay & Parameter  & &          &                     &            &   1.0 M  &            1.0 M \\
PR-BBB     & Generative replay & Parameter  & &          &                     &            &  20.2 M  &           20.2 M \\ \hline
\ourmodel  & Generative replay & Parameter  & &          &                     &            &   0.9 M  &            0.9 M \\
\Xhline{2.0\arrayrulewidth}
\end{tabular}%
}
\caption{Memory usage of replay-based methods on Split CIFAR-10}
\label{tab:memory_detail_cifar10}
\end{table}

\begin{table}[h!]
\centering
\renewcommand{\arraystretch}{1.15}
\resizebox{\textwidth}{!}{%
\begin{tabular}{lcccrrrrr}
\Xhline{2.0\arrayrulewidth}
& \multicolumn{1}{l}{} & \multicolumn{1}{l}{} & \multicolumn{1}{l}{} & \multicolumn{5}{c}{\textbf{Memory usage on Split   CIFAR-100}} \\ \cline{5-9} 
\multicolumn{1}{c}{Method} & Strategy  & Replay Target  &  & \multicolumn{1}{c}{Main model} & \multicolumn{1}{c}{Memory buffer} & \multicolumn{1}{c}{Generative model} & \multicolumn{1}{c}{Meta-model} & \multicolumn{1}{c}{Total} \\ \hline
A-GEM      & Memory replay     & Input      & &   11.2 M &    $\approx$ 7.7 M  &            &          &  $\approx$ 18.9 M \\
ER         & Memory replay     & Input      & &   11.2 M &    $\approx$ 7.7 M  &            &          &  $\approx$ 18.9 M \\
DGR        & Generative replay & Input      & &   11.2 M &                     &   13.8 M   &          &            25.0 M \\
BI-R       & Generative replay & Feature    & &          &                     &   13.3 M   &          &            13.3 M \\
HNET   & Generative replay & Parameter  & &          &                     &            &   5.0 M  &             5.0 M \\
PR-BBB     & Generative replay & Parameter  & &          &                     &            &  14.7 M  &            14.7 M \\ \hline
\ourmodel  & Generative replay & Parameter  & &          &                     &            &   5.1 M  &             5.1 M \\         
\Xhline{2.0\arrayrulewidth}
\end{tabular}%
}
\caption{Memory usage of replay-based methods on Split CIFAR-100}
\label{tab:memory_detail_cifar100}
\end{table}

\subsubsection{Experimental Details of \ourmodel}
The experimental details of our framework is presented in Table \ref{tab:exp_detail}. To ensure a fair comparison of learning abilities, we consistently use SGD with momentum as the optimizer when training the inference model in all the experiments.

\subsubsection{Performance and Memory Usage of Replay-Based Methods}
Table \ref{tab:app:exp_memory} summarizes the experimental results of our framework with \ourmodel and replay-based methods on CL benchmarks. The memory usage, strategies, and replay targets of each method are shown in Table \ref{tab:memory_detail_cifar10} for Split CIFAR-10 and Table \ref{tab:memory_detail_cifar100} for Split CIFAR-100. Memory usage in the tables is defined as the objects to be maintained for the next task learning, such as main inference models, memory buffer, generative models, and meta-models. The memory buffer has 100 samples per class. Assuming each pixel is 8 bits, the pixel occupies one-fourth the space of one parameter storage, which is a float tensor of 32 bits. Thus, one pixel is counted as 0.25 parameters.

\begin{figure*}[h!]
	\centering
    \subfigure[After learning the first\label{fig:chunk_visual:first} task]{\includegraphics[width=0.3\columnwidth]{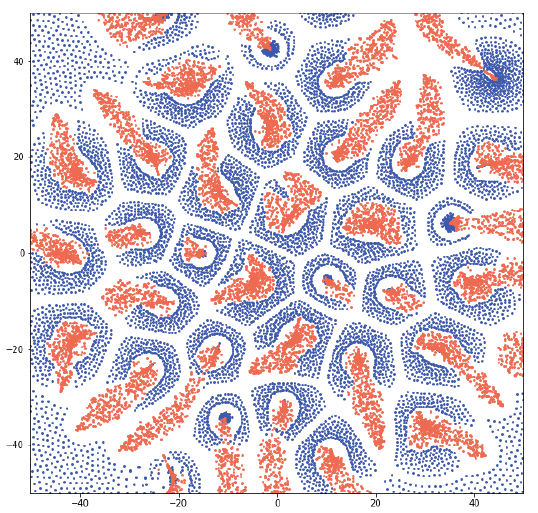}}
    \subfigure[After learning the last\label{fig:chunk_visual:last} task]{\includegraphics[width=0.3\columnwidth]{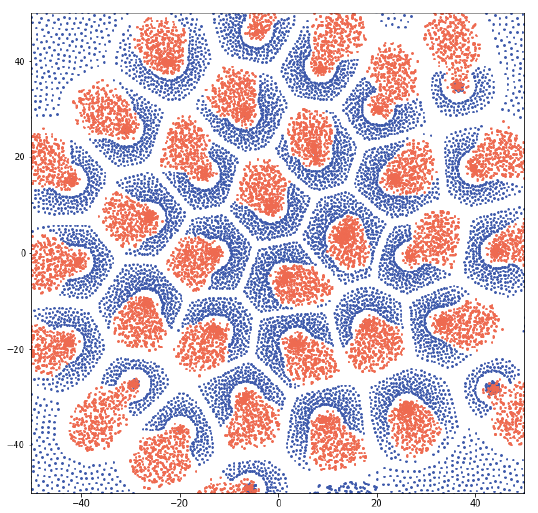}}
    \caption{(a) T-SNE visualizations of parameter chunks for the first task after learning the first task (a) and after learning the last task (b), respectively, on Split CIFAR-10 benchmark. The blue dots represent the sampled parameters from the BNN and the orange dots represent the generated parameters from the \ourmodel.}
    \label{fig:chunk_visual}
\end{figure*}

\begin{figure*}[h!]
	\centering
    \includegraphics[width=0.5\columnwidth]{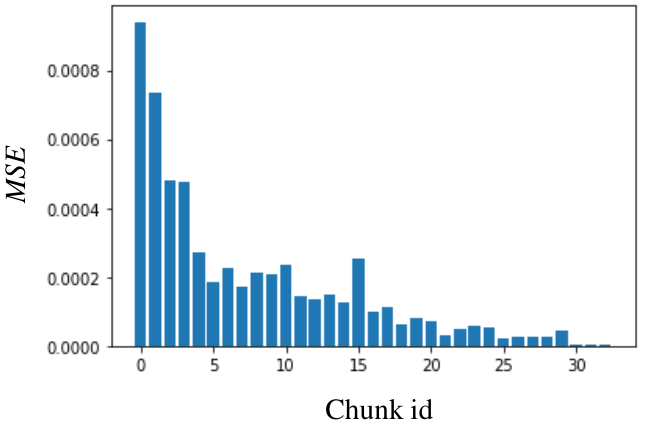}
    \caption{Mean squared error between the sampled parameters from the BNN and generated parameters from the \ourmodel on each chunk for the first task after learning the last task. MSE of individual parameters belonging to the same chunk is averaged.}\label{fig:chunk_mse}
\end{figure*}

\subsubsection{Visualizations of Generated Parameter Chunks}
In addition to measuring the accuracy of generated task-specific models $\theta^{\,t}_\G$ on CL benchmarks, we investigate the quality of parameter-generation. Figure \ref{fig:chunk_visual} visualizes how similar the generated chunks $\Phi_\G^{\,1}$ are distributed to the sampled chunks $\Phi_\B^{\,1}$ for the first task on Split CIFAR-10 benchmark. Figure~A\ref{fig:chunk_visual:first} is corresponding visualization after learning the first task and Figure~A\ref{fig:chunk_visual:last} is corresponding visualization after learning the final task. Each generated chunk from \ourmodel closely resembles the corresponding sampled chunk from the BNN and exhibits well-separated representation after learning the first task and still at the last task. We also compute mean squared error (MSE) between the generated parameters and the sampled parameters for the first task, $\text{MSE}(\theta^{1}_\G, \theta^{1}_\B)$, after learning the last task. Figure \ref{fig:chunk_mse} summarizes the results of mean squared error, where MSE of individual parameters belonging to the same chunk is averaged. For all chunks, the MSE values are less than 0.001.

%% file: main.bbl
\begin{thebibliography}{51}
\providecommand{\natexlab}[1]{#1}

\bibitem[{Ahn et~al.(2019)Ahn, Cha, Lee, and Moon}]{AhnCLM19}
Ahn, H.; Cha, S.; Lee, D.; and Moon, T. 2019.
\newblock Uncertainty-based Continual Learning with Adaptive Regularization.
\newblock In \emph{Advances in Neural Information Processing Systems}.

\bibitem[{Arjovsky, Chintala, and Bottou(2017)}]{arjovsky2017wasserstein}
Arjovsky, M.; Chintala, S.; and Bottou, L. 2017.
\newblock Wasserstein generative adversarial networks.
\newblock In \emph{International conference on machine learning}, 214--223. PMLR.

\bibitem[{Blundell et~al.(2015)Blundell, Cornebise, Kavukcuoglu, and Wierstra}]{blundell2015weight}
Blundell, C.; Cornebise, J.; Kavukcuoglu, K.; and Wierstra, D. 2015.
\newblock Weight uncertainty in neural network.
\newblock In \emph{International conference on machine learning}, 1613--1622. PMLR.

\bibitem[{Bulatov(2011)}]{bulatov2011notmnist}
Bulatov, Y. 2011.
\newblock Notmnist dataset. Google (Books/OCR).
\newblock Technical report, Tech. Rep.[Online]. Available: http://yaroslavvb. blogspot. it/2011/09~….

\bibitem[{Buzzega et~al.(2020)Buzzega, Boschini, Porrello, Abati, and Calderara}]{BuzzegaBPAC20}
Buzzega, P.; Boschini, M.; Porrello, A.; Abati, D.; and Calderara, S. 2020.
\newblock Dark Experience for General Continual Learning: a Strong, Simple Baseline.
\newblock In \emph{Annual Conference on Neural Information Processing Systems 2020, NeurIPS 2020}.

\bibitem[{Chaudhry et~al.(2019{\natexlab{a}})Chaudhry, Ranzato, Rohrbach, and Elhoseiny}]{ChaudhryRRE19}
Chaudhry, A.; Ranzato, M.; Rohrbach, M.; and Elhoseiny, M. 2019{\natexlab{a}}.
\newblock Efficient Lifelong Learning with {A-GEM}.
\newblock In \emph{International Conference on Learning Representations}.

\bibitem[{Chaudhry et~al.(2019{\natexlab{b}})Chaudhry, Rohrbach, Elhoseiny, Ajanthan, Dokania, Torr, and Ranzato}]{chaudhryER_2019}
Chaudhry, A.; Rohrbach, M.; Elhoseiny, M.; Ajanthan, T.; Dokania, P.~K.; Torr, P.~H.; and Ranzato, M. 2019{\natexlab{b}}.
\newblock Continual Learning with Tiny Episodic Memories.
\newblock \emph{arXiv preprint arXiv:1902.10486, 2019}.

\bibitem[{Davies and Bouldin(1979)}]{DaviesB79}
Davies, D.~L.; and Bouldin, D.~W. 1979.
\newblock A Cluster Separation Measure.
\newblock \emph{{IEEE} Trans. Pattern Anal. Mach. Intell.}, 1(2): 224--227.

\bibitem[{Ebrahimi et~al.(2020{\natexlab{a}})Ebrahimi, Elhoseiny, Darrell, and Rohrbach}]{EbrahimiEDR20}
Ebrahimi, S.; Elhoseiny, M.; Darrell, T.; and Rohrbach, M. 2020{\natexlab{a}}.
\newblock Uncertainty-guided Continual Learning with Bayesian Neural Networks.
\newblock In \emph{International Conference on Learning Representations}.

\bibitem[{Ebrahimi et~al.(2020{\natexlab{b}})Ebrahimi, Meier, Calandra, Darrell, and Rohrbach}]{ebrahimi2020adversarial}
Ebrahimi, S.; Meier, F.; Calandra, R.; Darrell, T.; and Rohrbach, M. 2020{\natexlab{b}}.
\newblock Adversarial continual learning.
\newblock In \emph{The European Conference on Computer Vision}, 386--402.

\bibitem[{Fisher(1936)}]{fisher1936use}
Fisher, R.~A. 1936.
\newblock The use of multiple measurements in taxonomic problems.
\newblock \emph{Annals of eugenics}, 7(2): 179--188.

\bibitem[{Goodfellow et~al.(2014)Goodfellow, Pouget-Abadie, Mirza, Xu, Warde-Farley, Ozair, Courville, and Bengio}]{NIPS2014_5ca3e9b1}
Goodfellow, I.; Pouget-Abadie, J.; Mirza, M.; Xu, B.; Warde-Farley, D.; Ozair, S.; Courville, A.; and Bengio, Y. 2014.
\newblock Generative Adversarial Nets.
\newblock In Ghahramani, Z.; Welling, M.; Cortes, C.; Lawrence, N.; and Weinberger, K., eds., \emph{Advances in Neural Information Processing Systems}, volume~27, 2672–2680.

\bibitem[{Gulrajani et~al.(2017)Gulrajani, Ahmed, Arjovsky, Dumoulin, and Courville}]{GulrajaniAADC17}
Gulrajani, I.; Ahmed, F.; Arjovsky, M.; Dumoulin, V.; and Courville, A.~C. 2017.
\newblock Improved Training of Wasserstein GANs.
\newblock In \emph{Advances in Neural Information Processing Systems}, 5767--5777.

\bibitem[{Ha, Dai, and Le(2017)}]{HaDL17}
Ha, D.; Dai, A.~M.; and Le, Q.~V. 2017.
\newblock HyperNetworks.
\newblock In \emph{International Conference on Learning Representations}.

\bibitem[{He et~al.(2016)He, Zhang, Ren, and Sun}]{HeZRS16}
He, K.; Zhang, X.; Ren, S.; and Sun, J. 2016.
\newblock Deep Residual Learning for Image Recognition.
\newblock In \emph{Conference on Computer Vision and Pattern Recognition}, 770--778.

\bibitem[{Henning et~al.(2021)Henning, Cervera, D'Angelo, von Oswald, Traber, Ehret, Kobayashi, Grewe, and Sacramento}]{HenningCDOTEKGS21}
Henning, C.; Cervera, M.~R.; D'Angelo, F.; von Oswald, J.; Traber, R.; Ehret, B.; Kobayashi, S.; Grewe, B.~F.; and Sacramento, J. 2021.
\newblock Posterior Meta-Replay for Continual Learning.
\newblock In \emph{Advances in Neural Information Processing Systems}.

\bibitem[{Izmailov et~al.(2018)Izmailov, Podoprikhin, Garipov, Vetrov, and Wilson}]{IzmailovPGVW18}
Izmailov, P.; Podoprikhin, D.; Garipov, T.; Vetrov, D.~P.; and Wilson, A.~G. 2018.
\newblock Averaging Weights Leads to Wider Optima and Better Generalization.
\newblock In \emph{Uncertainty in Artificial Intelligence}.

\bibitem[{Joseph and Balasubramanian(2020)}]{JosephB20}
Joseph, K.~J.; and Balasubramanian, V.~N. 2020.
\newblock Meta-Consolidation for Continual Learning.
\newblock In \emph{Advances in Neural Information Processing Systems}.

\bibitem[{Kemker and Kanan(2018)}]{KemkerK18}
Kemker, R.; and Kanan, C. 2018.
\newblock FearNet: Brain-Inspired Model for Incremental Learning.
\newblock In \emph{International Conference on Learning Representations}.

\bibitem[{Kingma and Welling(2013)}]{kingma2013auto}
Kingma, D.~P.; and Welling, M. 2013.
\newblock Auto-encoding variational bayes.
\newblock \emph{arXiv preprint arXiv:1312.6114}.

\bibitem[{Kirkpatrick et~al.(2017)Kirkpatrick, Pascanu, Rabinowitz, Veness, Desjardins, Rusu, Milan, Quan, Ramalho, Grabska-Barwinska et~al.}]{kirkpatrick2017overcoming}
Kirkpatrick, J.; Pascanu, R.; Rabinowitz, N.; Veness, J.; Desjardins, G.; Rusu, A.~A.; Milan, K.; Quan, J.; Ramalho, T.; Grabska-Barwinska, A.; et~al. 2017.
\newblock Overcoming catastrophic forgetting in neural networks.
\newblock \emph{Proceedings of the national academy of sciences}, 114(13): 3521--3526.

\bibitem[{Krizhevsky, Hinton et~al.(2009)}]{krizhevsky2009learning}
Krizhevsky, A.; Hinton, G.; et~al. 2009.
\newblock Learning multiple layers of features from tiny images.

\bibitem[{Krueger et~al.(2017)Krueger, Huang, Islam, Turner, Lacoste, and Courville}]{krueger2017bayesian}
Krueger, D.; Huang, C.-W.; Islam, R.; Turner, R.; Lacoste, A.; and Courville, A. 2017.
\newblock Bayesian hypernetworks.
\newblock \emph{arXiv preprint arXiv:1710.04759}.

\bibitem[{Lakshminarayanan, Pritzel, and Blundell(2017)}]{Lakshminarayanan17}
Lakshminarayanan, B.; Pritzel, A.; and Blundell, C. 2017.
\newblock Simple and Scalable Predictive Uncertainty Estimation using Deep Ensembles.
\newblock In \emph{Advances in Neural Information Processing Systems}.

\bibitem[{LeCun et~al.(1998)LeCun, Bottou, Bengio, and Haffner}]{LeCunBBH98}
LeCun, Y.; Bottou, L.; Bengio, Y.; and Haffner, P. 1998.
\newblock Gradient-based learning applied to document recognition.
\newblock \emph{Proc. {IEEE}}, 86(11): 2278--2324.

\bibitem[{Lopez-Paz and Ranzato(2017)}]{lopez2017gradient}
Lopez-Paz, D.; and Ranzato, M. 2017.
\newblock Gradient episodic memory for continual learning.
\newblock \emph{Advances in neural information processing systems}, 30.

\bibitem[{Lyu et~al.(2021)Lyu, Wang, Feng, Ye, Hu, and Wang}]{LyuW0YH021}
Lyu, F.; Wang, S.; Feng, W.; Ye, Z.; Hu, F.; and Wang, S. 2021.
\newblock Multi-Domain Multi-Task Rehearsal for Lifelong Learning.
\newblock In \emph{Thirty-Fifth {AAAI} Conference on Artificial Intelligence}.

\bibitem[{Maddox et~al.(2019)Maddox, Izmailov, Garipov, Vetrov, and Wilson}]{MaddoxIGVW19}
Maddox, W.~J.; Izmailov, P.; Garipov, T.; Vetrov, D.~P.; and Wilson, A.~G. 2019.
\newblock A Simple Baseline for Bayesian Uncertainty in Deep Learning.
\newblock In \emph{Advances in Neural Information Processing Systems}.

\bibitem[{Mallya and Lazebnik(2018)}]{mallya2018packnet}
Mallya, A.; and Lazebnik, S. 2018.
\newblock Packnet: Adding multiple tasks to a single network by iterative pruning.
\newblock In \emph{Proceedings of the IEEE conference on Computer Vision and Pattern Recognition}, 7765--7773.

\bibitem[{McClelland, McNaughton, and O'Reilly(1995)}]{mcclelland1995there}
McClelland, J.~L.; McNaughton, B.~L.; and O'Reilly, R.~C. 1995.
\newblock Why there are complementary learning systems in the hippocampus and neocortex: insights from the successes and failures of connectionist models of learning and memory.
\newblock \emph{Psychological review}, 102(3): 419.

\bibitem[{McCloskey and Cohen(1989)}]{mccloskey1989catastrophic}
McCloskey, M.; and Cohen, N.~J. 1989.
\newblock Catastrophic interference in connectionist networks: The sequential learning problem.
\newblock In \emph{Psychology of learning and motivation}, volume~24, 109--165. Elsevier.

\bibitem[{Netzer et~al.(2011)Netzer, Wang, Coates, Bissacco, Wu, and Ng}]{netzer2011reading}
Netzer, Y.; Wang, T.; Coates, A.; Bissacco, A.; Wu, B.; and Ng, A.~Y. 2011.
\newblock Reading digits in natural images with unsupervised feature learning.

\bibitem[{Ostapenko et~al.(2019)Ostapenko, Puscas, Klein, Jahnichen, and Nabi}]{ostapenko2019learning}
Ostapenko, O.; Puscas, M.; Klein, T.; Jahnichen, P.; and Nabi, M. 2019.
\newblock Learning to remember: A synaptic plasticity driven framework for continual learning.
\newblock In \emph{Proceedings of the IEEE/CVF conference on computer vision and pattern recognition}, 11321--11329.

\bibitem[{O’Reilly et~al.(2014)O’Reilly, Bhattacharyya, Howard, and Ketz}]{o2014complementary}
O’Reilly, R.~C.; Bhattacharyya, R.; Howard, M.~D.; and Ketz, N. 2014.
\newblock Complementary learning systems.
\newblock \emph{Cognitive science}, 38(6): 1229--1248.

\bibitem[{Parisi et~al.(2019)Parisi, Kemker, Part, Kanan, and Wermter}]{ParisiKPKW19}
Parisi, G.~I.; Kemker, R.; Part, J.~L.; Kanan, C.; and Wermter, S. 2019.
\newblock Continual lifelong learning with neural networks: {A} review.
\newblock \emph{Neural Networks}, 113: 54--71.

\bibitem[{Ratzlaff and Li(2019)}]{RatzlaffL19}
Ratzlaff, N.; and Li, F. 2019.
\newblock HyperGAN: {A} Generative Model for Diverse, Performant Neural Networks.
\newblock In \emph{Proceedings of the 36th International Conference on Machine Learning}, 5361--5369.

\bibitem[{Rebuffi et~al.(2017)Rebuffi, Kolesnikov, Sperl, and Lampert}]{rebuffi2017icarl}
Rebuffi, S.-A.; Kolesnikov, A.; Sperl, G.; and Lampert, C.~H. 2017.
\newblock icarl: Incremental classifier and representation learning.
\newblock In \emph{Proceedings of the IEEE conference on Computer Vision and Pattern Recognition}, 2001--2010.

\bibitem[{Riemer et~al.(2019)Riemer, Cases, Ajemian, Liu, Rish, Tu, and Tesauro}]{RiemerCALRTT19}
Riemer, M.; Cases, I.; Ajemian, R.; Liu, M.; Rish, I.; Tu, Y.; and Tesauro, G. 2019.
\newblock Learning to Learn without Forgetting by Maximizing Transfer and Minimizing Interference.
\newblock In \emph{International Conference on Learning Representations}.

\bibitem[{Saha, Garg, and Roy(2021)}]{SahaG021}
Saha, G.; Garg, I.; and Roy, K. 2021.
\newblock Gradient Projection Memory for Continual Learning.
\newblock In \emph{International Conference on Learning Representations}.

\bibitem[{Schacter and Addis(2007)}]{schacter2007cognitive}
Schacter, D.~L.; and Addis, D.~R. 2007.
\newblock The cognitive neuroscience of constructive memory: remembering the past and imagining the future.
\newblock \emph{Philosophical Transactions of the Royal Society B: Biological Sciences}, 362(1481): 773--786.

\bibitem[{Serr{\`{a}} et~al.(2018)Serr{\`{a}}, Suris, Miron, and Karatzoglou}]{SerraSMK18}
Serr{\`{a}}, J.; Suris, D.; Miron, M.; and Karatzoglou, A. 2018.
\newblock Overcoming Catastrophic Forgetting with Hard Attention to the Task.
\newblock In Dy, J.~G.; and Krause, A., eds., \emph{Proceedings of the International Conference on Machine Learning}.

\bibitem[{Shin et~al.(2017)Shin, Lee, Kim, and Kim}]{ShinLKK17}
Shin, H.; Lee, J.~K.; Kim, J.; and Kim, J. 2017.
\newblock Continual Learning with Deep Generative Replay.
\newblock In \emph{Advances in Neural Information Processing Systems}, 2990--2999.

\bibitem[{van~de Ven, Siegelmann, and Tolias(2020)}]{van2020brain}
van~de Ven, G.~M.; Siegelmann, H.~T.; and Tolias, A.~S. 2020.
\newblock Brain-inspired replay for continual learning with artificial neural networks.
\newblock \emph{Nature communications}, 11(1): 1--14.

\bibitem[{van~de Ven, Tuytelaars, and Tolias(2022)}]{van2022three}
van~de Ven, G.~M.; Tuytelaars, T.; and Tolias, A.~S. 2022.
\newblock Three types of incremental learning.
\newblock \emph{Nature Machine Intelligence}, 1--13.

\bibitem[{von Oswald et~al.(2020)von Oswald, Henning, Sacramento, and Grewe}]{OswaldHSG20}
von Oswald, J.; Henning, C.; Sacramento, J.; and Grewe, B.~F. 2020.
\newblock Continual learning with hypernetworks.
\newblock In \emph{International Conference on Learning Representations}.

\bibitem[{Wang et~al.(2018)Wang, Vicol, Lucas, Gu, Grosse, and Zemel}]{WangVLGGZ18}
Wang, K.; Vicol, P.; Lucas, J.; Gu, L.; Grosse, R.~B.; and Zemel, R.~S. 2018.
\newblock Adversarial Distillation of Bayesian Neural Network Posteriors.
\newblock In \emph{Proceedings of the 35th International Conference on Machine Learning}, 5177--5186.

\bibitem[{Wu et~al.(2018)Wu, Herranz, Liu, Wang, van~de Weijer, and Raducanu}]{WuHLWWR18}
Wu, C.; Herranz, L.; Liu, X.; Wang, Y.; van~de Weijer, J.; and Raducanu, B. 2018.
\newblock Memory Replay GANs: Learning to Generate New Categories without Forgetting.
\newblock In \emph{Advances in Neural Information Processing Systems}.

\bibitem[{Xiao, Rasul, and Vollgraf(2017)}]{xiao2017fashion}
Xiao, H.; Rasul, K.; and Vollgraf, R. 2017.
\newblock Fashion-mnist: a novel image dataset for benchmarking machine learning algorithms.
\newblock \emph{arXiv preprint arXiv:1708.07747}.

\bibitem[{Yoon et~al.(2018)Yoon, Yang, Lee, and Hwang}]{YoonYLH18}
Yoon, J.; Yang, E.; Lee, J.; and Hwang, S.~J. 2018.
\newblock Lifelong Learning with Dynamically Expandable Networks.
\newblock In \emph{International Conference on Learning Representations}.

\bibitem[{Zenke, Poole, and Ganguli(2017)}]{ZenkePG17}
Zenke, F.; Poole, B.; and Ganguli, S. 2017.
\newblock Continual Learning Through Synaptic Intelligence.
\newblock In \emph{Proceedings of the International Conference on Machine Learning}.

\bibitem[{Zhmoginov, Sandler, and Vladymyrov(2022)}]{Zhmoginov0V22}
Zhmoginov, A.; Sandler, M.; and Vladymyrov, M. 2022.
\newblock HyperTransformer: Model Generation for Supervised and Semi-Supervised Few-Shot Learning.
\newblock In \emph{International Conference on Machine Learning}, 27075--27098.

\end{thebibliography}
